\documentclass[journal,10pt]{IEEEtran}
\usepackage{amsmath}
\usepackage[utf8]{inputenc}
\usepackage{xcolor}
\usepackage{soul}
\usepackage{graphicx}
\usepackage[backend=biber,style=ieee]{biblatex}
\usepackage{multicol}
\def\lc{\left\lceil}   
\def\rc{\right\rceil}

\title{Dynamic Multi-objective Optimization of the Travelling Thief Problem}
\author{\IEEEauthorblockN{Daniel Herring\IEEEauthorrefmark{1}\IEEEauthorrefmark{2},
        Michael Kirley\IEEEauthorrefmark{2}, Xin Yao\IEEEauthorrefmark{1}}\\
\IEEEauthorblockA{\IEEEauthorrefmark{1}\textit{School of Computer Science, University of Birmingham, Birmingham, UK}\\
                  \IEEEauthorrefmark{2}\textit{School of Computing and Information Systems, University of Melbourne, Melbourne, Australia}}
      }
\begin{document}

\maketitle
\begin{abstract}
    \noindent Investigation of detailed and complex optimisation problem formulations that reflect realistic scenarios is a burgeoning field of research. A growing body of work exists for the Travelling Thief Problem, including multi-objective formulations and comparisons of exact and approximate methods to solve it. However, as many realistic scenarios are non-static in time, dynamic formulations have yet to be considered for the TTP. Definition of dynamics within three areas of the TTP problem are addressed; in the city locations, availability map and item values. Based on the elucidation of solution conservation between initial sets and obtained non-dominated sets, we define a range of initialisation mechanisms using solutions generated via solvers, greedily and randomly. These are then deployed to seed the population after a change and the performance in terms of hypervolume and spread is presented for comparison. Across a range of problems with varying TSP-component and KP-component sizes, we observe interesting trends in line with existing conclusions; there is little benefit to using randomisation as a strategy for initialisation of solution populations when the optimal TSP and KP component solutions can be exploited. Whilst these separate optima don't guarantee good TTP solutions, when combined, provide better initial performance and therefore in some examined instances, provides the best response to dynamic changes. A combined approach that mixes solution generation methods to provide a composite population in response to dynamic changes provides improved performance in some instances for the different dynamic TTP formulations. Potential for further development of a more cooperative combined method are realised to more cohesively exploit known information about the problems.
\end{abstract}

\section{Introduction}

Realistic formulation of existing and novel problems is increasingly becoming a focus for optimization research. This is particularly the case for the inclusion of features prevalent in real world scenarios, including the dynamics in complex tasks with multiple objectives. Much progress has been made in considering dynamics in both single objective problems \cite{Branke2002,Yang2013,Tinos2014,Rohlfshagen2013} and multi-objective optimization problems (DMOOPs) and algorithms with which to solve them \cite{Azzouz2016, Nguyen2012, Branke2003}. However, relatively fewer works address multi-objective combinatorial domains for this kind of problem with DMOTSP \cite{Yang2008,Yang2009} and DKP \cite{Farina2003} being the prevalent examples. Superposition of the static Travelling Salesman Problem (TSP) and Knapsack problems (KP) has been formulated as the Travelling Thief Problem; a realistic problem formulation with interconnected components \cite{Bonyadi2013}. Here, the proverbial ‘thief’ must simultaneously minimize their tour and maximize the profit of the items taken from the visited cities. The problem’s complexity and realism make it a good candidate problem to explore combinatorial DMOOPs.


Some electrical retailers in the UK, (e.g. John Lewis) offer an exchange service where old electrical appliances are collected when a delivery is made. Similarly, grocery delivery services (e.g. Ocado) offer services to take some recycling materials upon delivery. Furthermore, with the growth of the online food ordering industry (Uber Eats, Deliveroo), there is a capacity for these routing and item collection problems to be modelled as a dynamic TTP (DTTP) problem. The TTP provides a basis to model logistics and courier scenarios, however it may be a simplification to consider these problems as static. We therefore propose to take the first steps to include simple dynamics in the problem based on the dynamics seen in the dynamic TSP \cite{Psaraftis1988,Zhou2003} and dynamic Knapsack problems \cite{Kleywegt1998,Kleywegt2003}. 

There are many ways in which dynamics could be defined within the problem, however here we introduce three different types within the bi-objective TTP \cite{Bonyadi2013}. Since the TTP is effectively composed of a TSP with an overlaid KP, two of the types of dynamics correspond to changing city locations and item values. The third type is defined for the item availability map, denoting the items allocated to each city, as defined in the formulation of a TTP problem. Each of these is introduced in context of a generic courier collection problem to maintain the realism of the formulation (Section \ref{sec:Dynamics}).


The realistic nature of the TTP, together with it's composition from two well-known and extensively researched problems, has garnered much interest from the optimization and operational research communities. Many of the works aim to exploit the comprehensive existing research on each of the isolated TSP and KP components, including comparison of solver-based methods to an evolutionary algorithm \cite{Blank2017} or using solver-based methods to initialize algorithms applied to the TTP \cite{Faulkner2015, Wu2017}. 
However, the definition of the TTP is such that the TSP and KP components are interconnected, meaning that solving the TSP and KP parts in isolation does not guarantee good performance on the overall TTP \cite{Bonyadi2013} (further insights into interconnected problems can be found in \cite{Klamroth2017}). Despite this, it is intuitive that there is merit to providing an algorithm with good initial solutions to both TSP and KP components over random initialization in terms of finding good TTP solutions overall.


It is problematic however, to rely on exact solutions for the TSP and KP components to give good solutions to the TTP. Restricted exploration of search space regions that may yield good TTP solutions, despite sub-optimal performance in each separate component, is therefore a problem when using only solver-based or randomized starts.  Wagner notes in \cite{Wagner2016} for example, that good solutions for the TTP might require longer tours.


Investigating both the usefulness of domain knowledge incorporation (in the form of different initialization strategies) and its application to the dynamic formulation of the TTP are the main focuses of this work. We first determine the relative contributions of using different initialization methods for the static problem then extend their use to reactively seed the population in response to three types of dynamic changes in the DTTP with .


An important consideration is the instances on which we undertake this investigation. As Polyakovskiy \cite{Polyakovskiy2014} notes, EAs are suitable for TTP problems up to 3000-5000 cities and a format for different KP components is given in the competitions \cite{TTPcompGecco2019, TTPcompEMO2019, TTPcomp2017} that have been run for the single and bi-objective TTP (see Section 3). This range of problems adapted from TSPLIB \cite{Reinelt1991} provides a problem set with which existing methods for the bi-objective TTP have not been tested on; the work in \cite{Blank2017} used self-constructed problems and the largest problem instance used in \cite{Wu2018} and \cite{Yafrani2017} are $eil101$ and $kroA100$ respectively. Larger problems have been tackled for the single objective TTP \cite{Wagner2016, Mei2014} and it is noted that good solutions are common to both the single and bi-objective versions of the TTP \cite{Wu2018, Yafrani2017}.

Another vital point to address is the performance measurement in the DTTP; this is an important current research topic and a plethora of measurements for DMOEAs exist \cite{Helbig2013b, Camara2009}. However, as the exact Pareto Front (PF) for each problem is unknown reporting on the absolute performance of methods becomes difficult. Therefore we use both the hypervolume and Zitzler's spread \cite{Zitzler1999} (see Section \ref{sec:ExperimentsHypervolume}) to compare the relative performance of the applied response mechanisms.


To summarise, the contributions of this work are as follows. Firstly we observe the difference in localisation of solutions in the objective space based on a range of different initialisation strategies for the static TTP. Secondly, we extend the definition of the bi-objective TTP to include dynamics, considering  non-static environments in three intuitive and contextually meaningful aspects of the problem. These are city location change, item-city assignment changes (availability map) and item value changes. Thirdly, since exploitation of \textit{a priori} information for the TTP is a popular approach, combination of this with observed differences in performance between initialisation strategies, enables the construction and comparison of eight different responsive solution generation strategies for the dynamic TTP instances. Through this we examine the difference in the utility of introducing solutions generated via different strategies in enabling an evolutionary algorithm to continue to find competitive non-dominated sets across the dynamic intervals of the problem.



\section{Literature Review}
Dynamic multi-objective optimisation problems attempt to capture characteristics of realistic scenarios: the trade-off between multiple conflicting objectives and the non-static nature of the problem formulation. The basic formulation of which follows:\\

\begin{equation}
    \begin{aligned} 
      &\vec{\textbf{x}}=[x_{1}, x_{2}, \ldots x_{n}]\\
      \vec{\textbf{F}}(\vec{\textbf{x}},t)=&[f_{1}(\vec{\textbf{x}},t), f_{2}(\vec{\textbf{x}},t), \ldots f_{M}(\vec{\textbf{x}},t)]\\
      &h_{1}(\vec{\textbf{x}})\leq0, h_{2}(\vec{\textbf{x}})=0
    \end{aligned}
    \label{eqn:DMOOP}
\end{equation}
Equation \ref{eqn:DMOOP} gives one definition of a dynamic multi-objective problem with dynamics in the objective functions ($f(x,t)$), however alternative definitions can manifest the dynamics in the decision variables ($x(t)$), the number of decision variables ($n(t)$), the number of objectives ($M(t)$) or in the constraints ($h_{1}(x,t), h_{2}(x,t)$).
Extensive literature exists in studying the properties of DMOOPs \cite{Raquel2013}; on benchmarks \cite{Farina2003, Helbig2013a}, performance metrics \cite{Helbig2013b,Camara2009}, detection of changes \cite{Morrison2013} and algorithms to solve these types of problems \cite{Azzouz2017, Nguyen2012}. The unifying design purpose of algorithms for DMOOPs is to track the Pareto Front across the different dynamic environments in the problem and in a more efficient manner than random reinitialization of the population of solutions. Various strategies are employed to this end, including implicit and explicit memory techniques\cite{Branke2003}, diversity maintenance and introduction techniques \cite{Deb2007}, predictive methods (FPS and PPS \cite{Hatzakis2006, Zhou2014}) and hybrid methods \cite{Azzouz2017}. Some methods focus on generating solutions with fitnesses that are robust to change \cite{Yu2010}.

Fewer works specifically address DMOOPs in the combinatorial domain, with \cite{Farina2003} giving examples of simple dynamic multi-objective KP and TSP (DMO-KP and DMO-TSP resp.) benchmark formulation. Several works also exist for the DMO-TSP \cite{Yang2008,Yang2009}, for DMO scheduling problems \cite{Yang2008,Yang2014} and for the DMO Subset Sum problem \cite{Comsa2013}. However, no such consideration of dynamics in the Travelling Thief Problem exists.

The \textit{Travelling Thief Problem} is a realistic formulation of multi-component optimisation problems as a combination of interconnected TSP and KP components  \cite{Bonyadi2013}. As in a TSP, there is a number of cities, with a layout described by a distance matrix $D=\{ d_{ij}\}$, that must be ordered into the minimum distance Hamiltonian  \textit{tour}. Whilst on the tour, the `thief' in the problem must collect items from the cities according to a \textit{packing plan} that observes the capacity of a knapsack. Naturally, the objective functions of each component persist, with the goals being to minimize the tour length and maximize the profit of the items collected. 

The solution to the TSP component ($\bar{x}$) of the problem, a permutation of $N$ cities is evaluated via:\\
\begin{equation}
    f(\bar{x})=\sum_{i=1}^{n-1}(t_{x_{i},x_{i+1}})+t_{x_{n},x_{1}}, \bar{x}=(x_{1},\ldots,x_{n})
    \label{eqn:TSPtour}
\end{equation}
Where the travel time between adjacent cities $i$ and $i+1$.\\
\begin{equation}
    t_{x_{i},x_{i+1}}=\frac{d_{x_{i},x_{i+1}}}{v_{c}}
    \label{eqn:TSPtourtime}
\end{equation}
Where $d_{x_{i},x_{i+1}}$ is distance between cities $x_{i}$ and $x_{i+1}$ in the tour. The parameter $v_{c}$ represents the travel velocity. The optimal tour is the one with the minimum overall travel time.

Similarly, the KP component solution ($y$) is a combination of items, each of which has an associated profit (\textit{p$_i$}) and weight (\textit{w$_i$}). Each of the items selected contributes to the weight of the knapsack which must not exceed the capacity (\textit{W}). The optimal solution maximizes the total profit of the selected items whilst remaining within the capacity limit, calculated as:\\
\begin{equation}
    f(\vec{y})=\sum_{i\in \vec{y}}^{}p_{i} \quad | \quad \sum_{i\in \vec{y}}w_{i}<W
\end{equation}

To form the TTP, the knapsack problem is translated onto the TSP topology using an \textit{availability map} that describes which items can be selected from each city, to formulate a packing plan ($z$). Therefore a single candidate solution to the problem is comprised of a tour $x$ and a packing plan $z$ (which contains the knapsack solution information in $y$).

Despite these separate goals, solutions comprising the best tour and best knapsack solution may not provide the best TTP solutions. That is, solving each sub-problem in isolation does not provide information on the optimal solutions for the TTP. This interdependence of the problem is a key feature discussed in \cite{Bonyadi2013} and is an important characteristic of the TTP that was previously missing from existing benchmarks.

The authors define several methods by which the two sub-components can be connected: 
\begin{enumerate}
    \item Knapsack-usage dependent velocity.
    \item Time dependent item value degradation.
    \item Duplicate items with different values and locations in the availability map.
\end{enumerate}

As in the previously mentioned combinatorial DMOOP literature, the motivation for the incorporation of dynamics in these problems is to improve the realistic qualities of the formulation of a scenario. Some comparison can be drawn with the dynamic Pick-up and Delivery problem (DPDP) \cite{Saez2008}, however the key differences include the number of agents and agent’s policy towards the items.


There are two established types of DTSP within the literature in which different parts of the problem are considered non-static with time. The first type of DTSP occurs when the travel times between specific cities can change, initially designed to simulate traffic in the network of Dynamic Vehicle Routing Problems \cite{Psaraftis1988, Larsen2001, Bertsimas1990, Zhou2003}. The second kind of dynamics relates more generally to the cities\cite{Eyckelhof2002,Guntsch2001a,Guntsch2001b} whereby the number or locations of cities are the changing feature of the problem. Evolutionary approaches have been applied to both formulations \cite{Zhou2003, LishanKang2005}.

For the dynamic Knapsack problem (DKP), this is again an extension of the classical knapsack problem to include dynamic aspects of the problem that occur in real world scenarios, such as in transport logistics, batch processor scheduling \cite{Kleywegt1998} and bandwidth allocation problems \cite{Choi2002}. The dynamics can be in the number of knapsacks \cite{Perry2004}, or a change in the number, profits or weights of items \cite{Kleywegt1998, Kleywegt2003, Farina2003}.

Using these established DTSP and DKP dynamics together with the availability map generated in the definition of the TTP instance, we can formulate three versions of the DTTP, based on which type of dynamic behaviour the problem exhibits. Each of these can be explained in the context of a generic collection problem, e.g. a vehicle obtaining a list of items on a route (See Section \ref{sec:Dynamics}). The first uses a change in location of a number of cities (termed \textit{Loc}); the second type of dynamics operates as changes in the availability map (\textit{Ava}) and the third type of dynamics affects a change in the profits of the items (\textit{Val}).

The static TTP can be addressed as a bi-objective problem with objectives as shown in Eq \ref{eqn:TTP2obj}. \\
\begin{equation}
    G(x,z)=\begin{cases}
               min \quad f(x,z)\\
               max \quad g(x,z)
            \end{cases}
            \label{eqn:TTP2obj}
\end{equation}

where for $x$ and $z$, respectively the tour and packing plan, the function $f(x,z)$ is the travel time of the tour accounting for item selection; $g(x,z)$ is the value of items at the end of the tour, subject to their decaying value over the time (according to the Drop rate parameter) of the tour.

The value of $f(x,z)$ is calculated according to eqns. \ref{eqn:TSPtour} and \ref{eqn:TSPtourtime}, where $v_{c}$, the current speed of travel is calculated as:
\begin{equation}
    v_{c}=(v_{max}-W_{c}\frac{v_{max}-v_{min}}{W})
    \label{eqn:TTPvelocity}
\end{equation}
where $W_{c}$ and $W$ are the the current weight and maximum capacity of the thief's knapsack and $v_{min}$ and $v_{max}$ are the minimum and maximum velocity the thief can travel (fixed at 0.1 and 1). 

The value of $g(x,z)$ is the total value of the items at the end of the tour, calculated as:
\begin{equation}
    g(x,z)=\sum_{i \in z}^{}p_{i}*Dr^{\lc\frac{T_{i}}{C}\rc}
    \label{eqn:TTPprofit}
\end{equation}
where $Dr$ is the dropping rate, $T_{i}$ is the total time $i$ is carried during the tour and $C$ is a constant. The value of $Dr$ 0.9 and $C$ is calculated$C$ using the equation in~\cite{Bonyadi2013}, 
the the setting of the random value  on the interval $[0.2, 0.7]$ in this formula to $r=0.45$ so as to generate reproducible results. The calculation of $C$ is thus: 

\begin{equation}
    C=\frac{ln\left(Dr\right) * E_{t}}{v_{min} * ln \left(\frac{rl}{u}\right) }
\end{equation}
with the values of $l$ and $u$ being the minimum and maximum profit across all of the items. $E_{t}$ represents the shortest inter-city distance in the distance matrix.

As noted in the original bi-objective problem (TPP$_2$ in \cite{Bonyadi2013}), the first two of the connection methods mentioned above are included in the definition of the problem, the third is addressed in the different problem instances defined in \cite{Polyakovskiy2014} (Section \ref{sec:ExperimentsProblems}). 

Several recent works have sought to further understand the characteristics of the TTP and the factors controlling its difficulty. For example, the rent rate parameter has been studied and manipulated to generate more difficult instances of TTP problems \cite{Wu2016}. Fitness Landscape Analysis (FLA) has been conducted for the single objective TTP using Local Optima Networks for very small enumerable instances (7 cities, 6 items), confirming the intuitive notion that increasing the knapsack capacity (whilst maintaining all other instance settings) makes the problem easier to solve \cite{Yafrani2018c}. Similarly, Mei et al \cite{Mei2014} provides both theoretical and empirical research on the interconnectedness of the problem for the single objective case. Due to the impossibility of decomposing the objective function into additively separable components for the TSP and KP problems, the isolated solving of the sub-components followed by the combination of these to form solutions is less effective than consideration of the whole TTP problem, as set out in the foundational work \cite{Bonyadi2013}.
Some methods seek to use heuristics to solve the problem, such as simulated annealing and hill climbing \cite{Yafrani2018b} or evolutionary algorithms \cite{Wachter2015, Mei2016, Mei2014, Polyakovskiy2014, Blank2017} to find solutions for the TTP. Ant Colony Optimisation \cite{Birkedal2015, Wagner2016}, Co-operative Coevolution \cite{Mei2014, Bonyadi2014} and Local Search \cite{Polyakovskiy2014} methods have also been applied to the single objective TTP. More information on all these algorithms can be found in \cite{Wagner2017}.

A popular methodology relies on incorporating exact information from deterministic methods. Faulkner et al. \cite{Faulkner2015} defines some approximate heuristics methods for solving the single objective TTP that are initialised with tours using the Chained Lin-Kernighan TSP solver \cite{Applegate2003}. Exact methods have also been employed in \cite{Wu2017} to determine the performance, by comparison, to a range of approximate methods applied to TTP problems. Also defined here are hybrid methods that combine Dynamic Programming (DP) with the S5 heuristic from \cite{Faulkner2015} to find the optimal packing plan to the Packing While Travelling (PWT) problem \cite{Polyakovskiy2015, Polyakovskiy2017} (fixed tour scenario of the TTP). A similar use of DP for PWT is combined in a hybrid method for the bi-objective TTP problem \cite{Wu2018}. 

Relatively few works address the bi-objective version of the problem \cite{Yafrani2017, Blank2017, Wu2018}. Blank et al’s approach uses solvers combined with low level heuristics to provide solutions to a limited set of TTP instances \cite{Blank2017}; however it remains to be seen whether these findings translate to the comprehensive benchmark problems \cite{Polyakovskiy2014} proposed for the TTP (constructed on TSPLIB). The work in \cite{Blank2017} also neglects the \textit{Drop} in item value feature of the multi-objective definition of the problem \cite{Bonyadi2013}.
As noted previously, Polyakovskiy et al, \cite{Polyakovskiy2014} state that EAs provide competitive results on TTPs with up to 3000-5000 cities, however the range of problems tested in the aforementioned literature for EAs is largely restricted to much smaller problems. The exception to this can be found in the work of Mei et al \cite{Mei2014}, which looked specifically at large scale problems with at least 10,000 cities and 1,000,000 items. The MMAS approach (ACO based) in \cite{Wagner2016} is applied on instances with at most 1000 cities and 10,000 items and contains the greatest range of problems that any method for the TTP is tested on. It is worth noting that of the studies that consider specifically the bi-objective TTP problem, the base problems used are limited to \textit{eil51, eil76, eil101} \cite{Wu2018} and \textit{kroA100} \cite{Yafrani2017}.
Of these studies concerned with the bi-objective version of the problem, an EA has been used in both and it is noted that the single-objective problem is contained within the bi-objective one, such that good solutions are common to both \cite{Yafrani2017}. The combination of performance measures used allow for inference of the characteristics of the achieved solution set in a multi-objective space more intuitively than for the single objective case.

Use of known good solutions to help solve TSP instances \cite{Oman2001} and the seeding of populations both at initialisation \cite{Johnson2005,Kazimipour2014,Friedrich2015,Dasgupta2009} and during the optimisation of dynamic MOOPs \cite{Hatzakis2006, Zhou2007, Zhou2014} have been shown to improve the rate of convergence to the Pareto Front (PF) to varying degrees. Therefore, initialisation strategies that incorporate exact knowledge in the form of known good solutions to the different TSP and KP components of the TTP (e.g. as in \cite{Faulkner2015}) are likely to provide some good solutions and a better performing starting population than a random initialisation strategy. It is noted that seeding 100\% of the population with known good solutions is not always helpful \cite{Keedwell2005, Oman2001}. Additionally, as noted previously in \cite{Wagner2016}, better solutions to the TTP may require longer TSP tours and as such a mixed approach to initialisation is likely to improve convergence further. For example, the perturbation of the solver-based TSP tours and KP packing plans and the inclusion of greedily constructed tours and packing plans.

In this work we keep the mixed mutation strategy of Blank et al \cite{Blank2017}, but instead of using the solvers and the EA separately, we use the different methods (solver, greedy and random) to initialise populations that are passed to a generic EA framework, as starting points for the evolution. Preliminary examination in a static TTP environment allows inference of the importance of initialization strategies for the dynamic problem. For the dynamic TTP instances, the population generation methods are deployed in response to dynamic changes in the problem to seed the population with solutions. A dynamic version of the TTP, has to the best of the authors' knowledge, not been proposed and herein we draw from the dynamics seen in both dynamic TSP and dynamic KP literature in order to formulate the DTTP.

In both \cite{Blank2017} and \cite{Wu2018}, the hypervolume is used as a performance measurement for the bi-objective TTP, with an unknown reference point and [0,\textit{knapsackCapacity}] used respectively, despite the second objective relating to the profit of the items rather than their weight. Modified versions of the hypervolume metric have also been used for DMOOPs (a summary of which is included in \cite{Helbig2013b}), to describe the stability and accuracy of the population of solutions across dynamic intervals \cite{Camara2009}.  In this respect, for problems with unknown Pareto Fronts, hypervolume-based metrics together with a measure of the spread of the achieved solution set enable comparative performance analysis between the different response methods.

\section{Methods}

The types of dynamics present in dynamic optimisation problems varies greatly across the literature. Commonly, a DMOOP has a time dependency within it's objective functions, however there are many other parts of problems that can host dynamics. These include as previously described: the decision variables and the number of them; the number of objectives; or within the constraints of the problem. 
Problems with dynamic features can be classified in a number of ways, including  by their magnitude, frequency or regularity (whether the dynamics are cyclic or repeating) \cite{Azzouz2016}. Problems have also been classified by the effect of the dynamics on the Pareto Set and Pareto Front \cite{Farina2003}.

\subsection{Dynamic TTP Characteristics}\label{sec:Dynamics}
Within the TTP there are several aspects of the problem which can be modified to reflect the dynamics we see in real systems. These include the node (city) locations, item availability and the item values. The dynamics introduced to each of these aspects will be explained in context below as a modification to the standard bi-objective TTP problem. To enable a greater understanding of the impacts these different dynamics have on the problem, within this work each dynamic modification is studied in isolation.

A solution to the TTP takes the form of a tour ($x$) and a packing plan ($z$). The time taken to complete the tour component is evaluated according to the objective functions in Eq.\ref{eqn:TTP2obj} using a symmetrical distance matrix, $D$, of Euclidean distances between every pair of cities. The packing plan portion of the solution is comprised of a sequence of items that are collected from specific cities during the tour; the location of each item being determined by the problem-specific availability map ($A$). As with classical KP, each item ($I$) has a weight ($I_{w}$) and a profit ($I_{v}$) associated with it.

\subsubsection{Dynamic City Location}\label{sec:DynamicsLoc}

Within the DTSP literature, dynamics are added in the form of changing locations of the cities, adding or removing cities or altering specific distances between cities to replicate traffic in the network. We focus on the first of these for the TTP, opting to use a context-driven justification for the dynamics proposed.

Where a city ($n$) is represented with Cartesian coordinates ($n_{x}, n_{y}$), the distance matrix, $D$ contains the Euclidean distance between each pair of cities: $d(i,j)= \sqrt{({n_{i,x}-n_{j,x})}^2 + {({n_{i,y}-n_{j,y})}^2}}$ for $i,j=1,...,N$. Changing the location of the city requires the alteration of the city coordinates and the re-calculation of the row and column elements of $D$ that relate to the chosen city. To determine the allowable translation limits for a city's location, the initial range in $x$ and $y$ directions is increased by 5\% in each direction, giving a maximum of a 10\% total increase in length along each axis (whilst maintaining non-negative coordinate values). The new coordinates are calculated for the next dynamic interval in Eqn. \ref{eqn:DynamicsLoc}:

\begin{equation}
    n_{x,i,t+1} = r_{1}, \quad n_{y,i,t+1} = r_{2}
    \label{eqn:DynamicsLoc}
\end{equation}
where $r_{1}, r_{2}$ are random numbers drawn from the uniform discrete distributions $unif_{1}\{\min(\min_{i=1,...,N}(n_{x,i,0}),0), \max_{i=1,...,N}(n_{x,i,0})\}$ and $unif_{2}\{\min(\min_{i=1,...,N}(n_{y,i,0}),0), \max_{i=1,...,N}(n_{y,i,0})\}$ respectively.
A number of cities ($d_{N}$ - representing the magnitude of the change) have their coordinates updated to randomly generated locations within this feasible area and the corresponding entries in the distance matrix are updated to transition to the next dynamic interval of the problem.

In the context of a realistic scenario, given a courier that must collect items along it's route subject to the objective function definitions as before, a change in a city location can be interpreted as an alternative depot (with the same items) being chosen. Some example motivations for this could be a closure or road incident preventing access to the original location.

\subsubsection{Dynamic Item Availability}\label{sec:DynamicsAva}

The availability map of a TTP problem denotes which items are located in which cities and is defined together with the initial city locations and item weight and profit distributions. Since this is a design feature of the TTP, there is no precedent for dynamics within this aspect of the problem and so we propose the following: a dynamic change in the availability map corresponds to a change in item-city assignments ($I_{city}$) for a number of items. Since the range of TTP instances can have different numbers of items, we propose that the magnitude of change be controlled as a percentage of the total items (as $d_{N}$) which undergo an assignment change. The formula for a change in an item's city assignment can be written as:

\begin{equation}
    I_{city,t+1} = r
\end{equation}
where $r$ is a random city index drawn from the uniform discrete distribution $unif\{1,N\}$. It is possible that an item's new city index is the same as its previous city index. As with the changing city location, a change in the availability map can be contextually interpreted as stock shortages or discontinued items at the item's original city index, and thus a switch to an alternative depot location (city) or a competitor offering the same items.

\subsubsection{Dynamic Item Values}\label{sec:DynamicsVal}

Dynamic Knapsack problems, as mentioned previously can be defined to have non-static capacities or numbers of items, their weights or profits. In the context of a courier collection problem, only the non-static number of items and item profit make logical sense, however to change the number of items also requires updating the availability map and so we limit this form of the dynamics to the item profits only. As with the availability map dynamics, the magnitude of each change is determined by both the number of item profits that change but also by how much they change. The percentage $d_{N}$ gives us the number of items that change, and the \textit{change factor, $cf$} give the amount of change in each item's profit. An item's profit value is updated as:

\begin{equation}
    I_{p,t+1} =  (1 + (s \times cf)) \times I_{p,t}  
\end{equation}
where $s$ controls the sign of the change, chosen (uniformly) at random for each item. The profit of items in a realistic setting will always be changing due to a number of factors. These include, global and local stock levels, competition factors, inflation and in terms of perishable items, this can relate to the freshness or quality of the items.

\subsection{Problems}\label{sec:ProblemsDynamics}
DMOOPs have several important characteristics in terms of their dynamics: the magnitude of changes, the frequency of changes and the regularity or cyclicity of changes. In terms of the magnitude, as described for each of the types of dynamics this is regulated by the parameter $d_{N}$ and for the \textit{change factor} in dynamic item values, we fixed this at 2\%. To enable clear comparison between strategies, the frequency of change is fixed at 200 generations and $d_{N,Ava/Val}=5\%$ and $d_{N,Loc}=2$ parameters are used.

\subsubsection{Reproducible Dynamic Instances} \label{sec:ProblemsReproducible}
In order to make the dynamic changes reproducible to test the different solution response methods a number of seed files were generated. These contained pre-generated uniform random numbers that would enable reproducible dynamic events to occur such that the different algorithm methods were exposed to the same problem instances exactly. This process also means that there is no deliberate cyclicity to the dynamic changes that occur.

\subsection{Algorithm and response mechanisms}\label{sec:Algorithms}

As described previously, an individual solution to the TTP is comprised of a tour $\pi$ (a permutation of the cities) and a packing plan $z$ (a selection of chosen items and the cities they are collected from). Thus the process of seeding the initial population with good solutions must address both of these solution components separately. If we consider firstly, the static bi-objective TTP, we can examine the three groups of initialization strategies in terms of their contributions to expediting convergence. These three groups are solver-based, greedy and random initialization methods.

\subsubsection{Evolutionary Algorithm}\label{sec:AlgorithmsEA}
\begin{figure*}[!t]
    \centering
    \includegraphics[width=\textwidth]{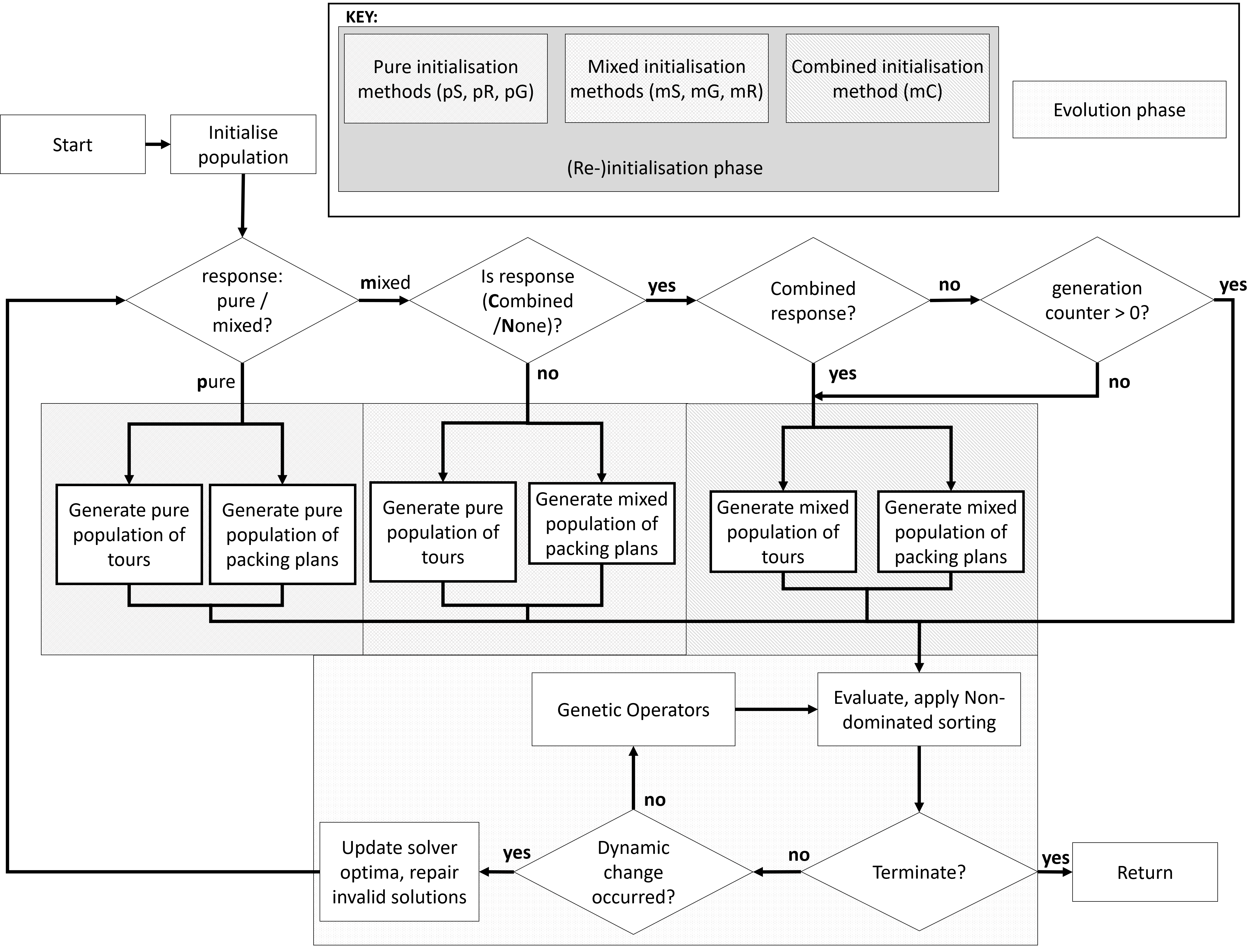}
    \caption{Illustration of algorithm operation for multiple solution generation strategies for the DTTP. Where a solution is generated using a solver or greedily, a mutation based perturbation is applied to this solution to meet the quota of the population required from this method. In reference to `pure' and `mixed' methods, these correspond to respectively using one or all solution generation methods from \{solver, random, greedy\}. For the DTTP, the distinct response strategies are explicitly summarised in Table \ref{tab:Algorithms}.}
    \label{fig:AlgFlowchart}
\end{figure*}

The evolutionary algorithm framework employed combines the genetic operators as applied in \cite{Blank2017}, with the popular non-dominated sorting approach to replacement of solutions from NSGA-II \cite{Deb2002}. The structure of the algorithm is shown in Figure \ref{fig:AlgFlowchart} and is comprised of two phases: the response phase and the evolution phase. Depending on the initialisation method (ergo the response strategy in the DTTP case), the construction of the population and the individual solutions themselves differs. More information about these different strategies is given below.

\subsubsection{Solver-based initialization}\label{sec:AlgorithmsSolver} 
There exist several exact solvers for TSP problems, for example Concorde \cite{Applegate1998} and Branch and Bound methods \cite{Little1963}. To remain consistent with the methods in \cite{Blank2017}, we use the  Lin-Kernighan heuristic (LKH v2.0.9, \cite{Helsgaun2000}) to provide the optimal tour component. For solver-based initial solution for the packing plan, the KP problem is solved via a simple dynamic programming (DP) approach, (however this becomes too computationally demanding for some of the problem instances examined - see Section \ref{sec:AlgorithmsSolver}). The optimal KP solution is then transformed into the packing plan format using each item's city index in the availability map. These two approaches give us a single exact optimal solution for each component and so to avoid the issues with over-seeding, we employ the mutation operator for each solution component; Bitflip for KP solutions, (before conversion to packing plan) and Single Swap Mutation for the TSP tour, until we reach a number of unique solutions equal to the population size.

\subsubsection{Greedy Initialization}\label{sec:AlgorithmsGreedy}
Greedy initialization is a well known strategy that in some cases can provide near-optimal solutions and in others is little better than randomized solutions. For the tour component of a solution, since we fix our tours to begin at the first city (city index 1, it has no items) we select the shortest distance to any city from here and add this to the tour, then repeat this process from the newly added city (ties are broken by lower city index value), this process repeats until a complete tour has been established.

For the KP solution, the items are sorted in descending order of their profit/weight ratio and the all items with a cumulative weight below the knapsack capacity $C$ comprise the greedy solution. Similarly to the solver-based tours, the single solution in mutated using the algorithm operators to form an initial population. 

\subsubsection{Random Initialization}\label{sec:AlgorithmsRandom}
The randomly initialized population is comprised of random permutations for the tour and a random permutation of items, truncated at the point where their cumulative weight exceeds the maximum capacity of the knapsack. This is then converted to a packing plan as before.

\subsection{Re-initialization for the DTTP} \label{sec:MethodsRein}


For any MOEA method applied to DMOOPs, as long as the performance of the algorithm is better than is achievable compared to random re-initialization, then the algorithm can be described as useful. Therefore, given the two-component nature of the TTP solutions and the intuitively different information gained using solvers, greedy and random initializations, testing a combination of these methods will prove helpful in ascertaining the impact of solution choice for population seeding in the dynamic TTP. 

Preliminary results indicate the presence of differences in the solutions obtained when using different initialization for the tour and packing plan components.

Therefore, we propose a collection of eight different strategies, each injecting solutions into the population using different combinations of initialization methods, in response to a change in the problem environment. These are illustrated through the various routes in Fig \ref{fig:AlgFlowchart} and are summarised in Table \ref{tab:Algorithms}.

\begin{table*}[!t]
\begin{center}
\begin{tabular}{c|c|cll}
\textbf{Response Strategy} & \textbf{TSP solution} & \textbf{Packing Plan} &  &  \\ \cline{1-3}
pS                 & solver                & solver               &  &  \\
pG                 & greedy                & greedy               &  &  \\
pR                 & random                & random               &  &  \\
mS                 & solver                & solver/greedy/random &  &  \\
mG                 & greedy                & solver/greedy/random &  &  \\
mR                 & random                & solver/greedy/random &  &  \\
mC                 & solver/greedy/random  & solver/greedy/random &  &  \\
mN                 & none*                 & none*                &  & 
\end{tabular}
\end{center}
\caption{Re-initialization strategies for TSP and Packing Plan solution components employed by the different strategies applied to the DTTP instances. The mN algorithm uses the mC initialization but has no response mechanism to dynamic changes.}
\label{tab:Algorithms}
\end{table*}
Since the goal of this work is to understand the impact of solution information, detection of the changes is not considered by the algorithm, instead the dynamic response is triggered according to a fixed schedule of changes (see section \ref{sec:ProblemsReproducible})

The performance of algorithms improves according to the standard measures for static problems, i.e. convergence quality and speed, spread and extent of the solutions found as measured by a variety of metrics. To provide general information on these qualities of the solutions sets, we employ the widely used hypervolume and spread \cite{Zitzler1999} metrics.

\subsection{Experimental setup}
To understand how the use of these different methods of solution generation allow for mitigation of the impacts of dynamic changes to the TTP, we examine their performance on a variety of scenarios. Since each initialization/response strategy provides the optimisation algorithm with a population of solutions with different focuses, e.g. randomized solutions increase diversity in the population, whilst solver-based solutions in each component, will, intuitively skew the populations towards optimal values in their corresponding objective. 

Therefore, to ascertain the merit of exploiting known information for the TSP and KP components of the problem and using this to find good solutions to the TTP as whole, we select TTP instances in accordance with the the competition format of TTP problems \cite{TTPcompGecco2019, TTPcompEMO2019, TTPcomp2017}, as mentioned previously.

Using this format of experiments and measurements, we hope to examine the effect that providing differently constructed solutions has in enabling continued success of an algorithm in the post-change problem environment.

\subsubsection{Problems}
\label{sec:ExperimentsProblems}

    For a TTP problem with each type of dynamics, 10 dynamic instances are constructed, each of which comprises a different set of fixed changes. The changes themselves are generated (reproducibly, see Section \ref{sec:ProblemsReproducible}) uniformly at random. For each of these instances 30 repeats are performed with each response method.
    
\begin{figure*}[!t]
    \centering
    \includegraphics[width=\textwidth]{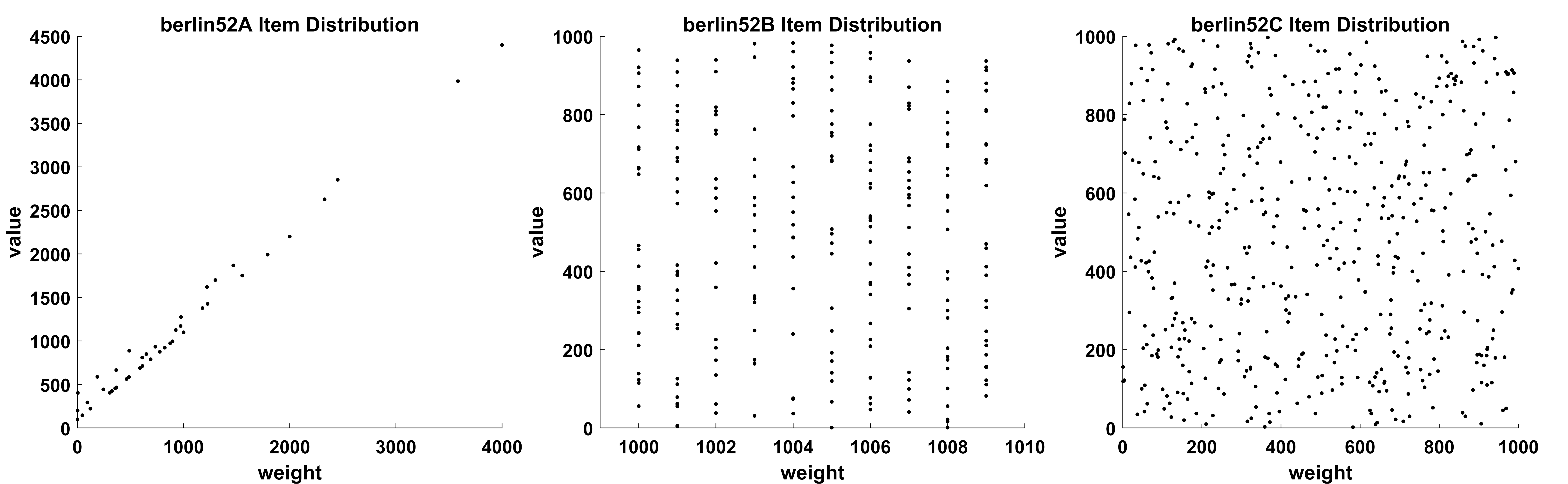}
    \caption{Item weight vs. value distributions using berlin52 TSP problem with A, B \& C knapsack problem types.}
    \label{fig:ABCitemdists}
\end{figure*}

    The base TTP problems are modified to contain the dynamics and follow the aforementioned TTP competition format. We use 4 different sizes of TSP components (berlin52, a280, rat783, u2319) and for each of these there are three problem types (A, B, C) that alter the knapsack and TTP-specific properties with different numbers of items per city and different item weight/value ratios. The types differ in their knapsack capacities as well as in their distributions of item weights and profits as shown in Fig. \ref{fig:ABCitemdists}. They can be described as follows:
    
    \begin{itemize}
        \item A: 1 item per city, lowest knapsack capacity, strongly correlated item weights/profits.
        \item B: 5 items per city, medium knapsack capacity, similar item weights/profits.
        \item C: 10 items per city, highest knapsack capacity, uncorrelated item weights/profits.
    \end{itemize}

\subsubsection{Measurements}
\label{sec:ExperimentsHypervolume}
In order to facilitate clearer comparison between the different strategies, we set the nadir point to be ($f^{\dagger}_{tour},f^{\dagger}_{profit}$)=$[\overline{D}\times|D|, 0]$. Any solution set that does not dominate this point is given a hypervolume value of 0. This means that all problem instances with a common TSP component will share a reference point before any changes occur.

\label{sec:ExperimentsPlotDescriptions}

The values of the hypervolume and spread are averaged over their repeats for each dynamic instance (pattern of changes) and then for the five dynamic intervals in a dynamic instance, the end-of interval hypervolume is ranked across the response mechanisms. These ranks are aggregated across the instances and the median of these ranks is then presented in  polar plots. The presented Hypervolume and spread profiles illustrate the mean across the dynamic instances for these measurements.

\section{Results}

\subsection{Solution localization within the non-dominated set based on initialization (Static TTP problem)}
\begin{figure*}
    \centering
    \includegraphics[width=\textwidth]{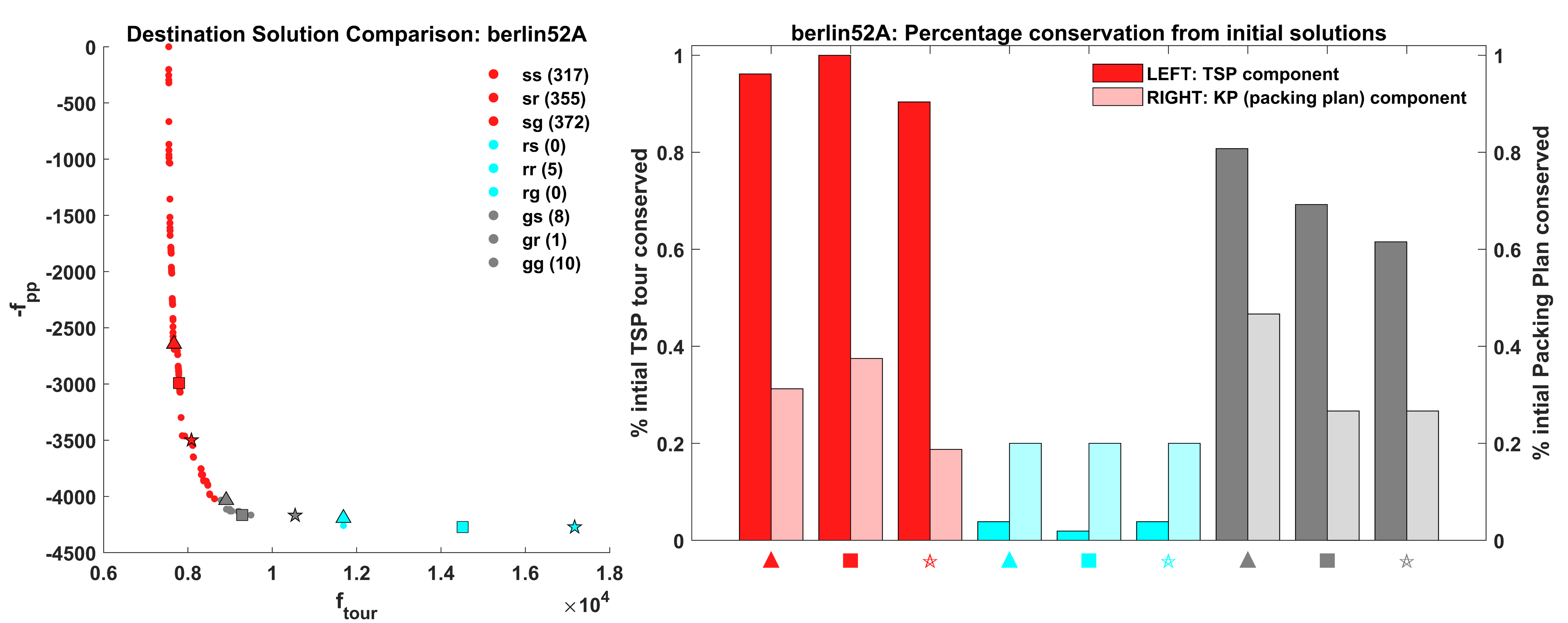}
    \caption{(left) The composite Pareto Front obtained for the berlin52A problem using the different initialization strategies. Each algorithm ran for 1000 iterations, with a population size of 60. The legend entries correspond to the strategies used for initialization; `gg' for example refers to the algorithm initialized with greedily formed TSP and KP solutions and the number in parentheses denotes the number of contributed  non-dominated solutions in the composite PF. (right) The percentage conservation from initial solution components to representative solutions in the non-dominated set are given, colours and marker shapes correspond to those used in the left-hand plot.}
    \label{fig:SC_berlin52A_pure_joined}
\end{figure*}

Preliminary investigation of the three types of initialization techniques on the static problem as they are described previously reveals interesting insights. The localization of the final population of solutions appears dependent on the method used to initialize the population. For example in Fig. \ref{fig:SC_berlin52A_pure_joined} we see a composite non-dominated set using the aggregation of final populations of solutions achieved when using different `pure' initialization methods. The term $pure$ means that the same strategy is used for both TSP and KP initial solution components (e.g. $ss$, TSP-solver \& KP-solver). 

We can clearly see that different regions of the non-dominated set are represented by the strategies initialized using the different methods. Figure \ref{fig:SC_berlin52A_pure_joined} shows clearly that using solvers for both solution components (tour and packing plans) results in many unique contributions to the non-dominated set being found with short tours. In contrast, the greedily initialised solutions obtain higher profit solutions with slightly longer tours and the randomly initialized solutions lead the algorithm to some of the highest profit solutions. To examine the causes of this localised convergence we calculate two comparative measurements on the initial and final solutions based on the percentage of the initial tour and packing plan that is conserved within the solutions. Intuitively, this initial solution for solver and greedy based methods is a single solution. In the case of random methods, the same procedure was followed in that a single tour and packing plan are generated and perturbed using operators to fill the population with unique solutions. The justification for this is that it provides a comparable and illustrative example for solution conservation, however it is obvious that a population of solutions generated randomly and entirely independently from each other will have more diversity and could perform better, therefore this approach is adopted in subsequent experiments.

To calculate the conservation from the tour component of the initial solutions, the percentage of neighbouring pairs of cities that are retained in the final solutions are compared with their occurrence in the initial solution. For the packing plan, the percentage is a composition of the percentage of item indexes that match between initial and final solution chromosomes and the percentage similarity of cities where these items are collected. (e.g. 100\% conservation is obtained by same items from the same cities. 50\% conservation can be obtained by different items from exactly the same cities).

Clearly visible in Fig.\ref{fig:SC_berlin52A_pure_joined} we see that the labelled solutions for solver-based initialization strategy (herein, $ss$) retain at least 90\% of their initial (optimal) tour and less that 40\% of their initial packing plan. This enables the algorithm to pursue the minimum $f_{tour}$ solutions since these solutions are already competitive and the optimisation process for this algorithm is mostly exploring alternative packing plans. The labelled solutions from contributed by the greedily-initialized strategy (herein, $gg$), may not have (depending on the specific problem) optimal TSP-components upon initialization and therefore the possibility for exploration is greater since improvements can be made by simultaneous exploration of tours and packing plans, whereas for $ss$, most alterations to the tour-component will have resulted in a poorer fitness in $f_{tour}$ and thus not being retained for future generations. This is illustrated in the conservation of the initial tours and packing plans for the $gg$ solutions. We see less initial  tour (greedy - non-optimal) conservation than compared with the $ss$ solutions and similar levels of packing plan conservation between the solutions from the two strategies.

The solutions contributed by the randomly initialized algorithm (herein, $rr$) have the highest profits and also the highest tour lengths. For these solutions, we see very little conservation of the initial tours (random permutations likely have very poor fitness), together with some minor conservation of the initial packing plans. The poor initial $f_{tour}$ fitness for solutions in this algorithm mean that, as with the greedy approach, exploration is not limited by the already competitive solutions present in the population. This enables exploration of the highest profit solutions since progress is not already partially driven towards lower $f_{tour}$ solutions as in the other two strategies.

This illustrates the observation in \cite{Wagner2016}, which notes that longer tours are necessary to find good TTP solutions; the greedy- and randomly-initialized strategies find non-dominated solutions with longer-than-optimal tour lengths.

Another important observation from this is in the proportion of initial information that is being exploited and the effect that it has on the observably localized convergence of the algorithm. Optimal/good TSP solution information is highly preserved during the evolution thanks to the directness of its impact on the solution fitness. The optimal/good KP solution information supplied sees markedly less conservation, likely due to the indirectness of an optimal KP solution on the $f_{pp}$ fitness, since it must first be transformed into a packing plan, which imposes an order on the KP solution that is not inherent in its formulation. Ideally, given the optimal tour, we could generate an optimal packing plan for this tour, known as the Packing While Travelling problem (PWT), however this remains an NP-hard problem in itself \cite{Polyakovskiy2015, Polyakovskiy2017}.

\begin{figure*}[!t]
    \centering
    \includegraphics[width=\textwidth]{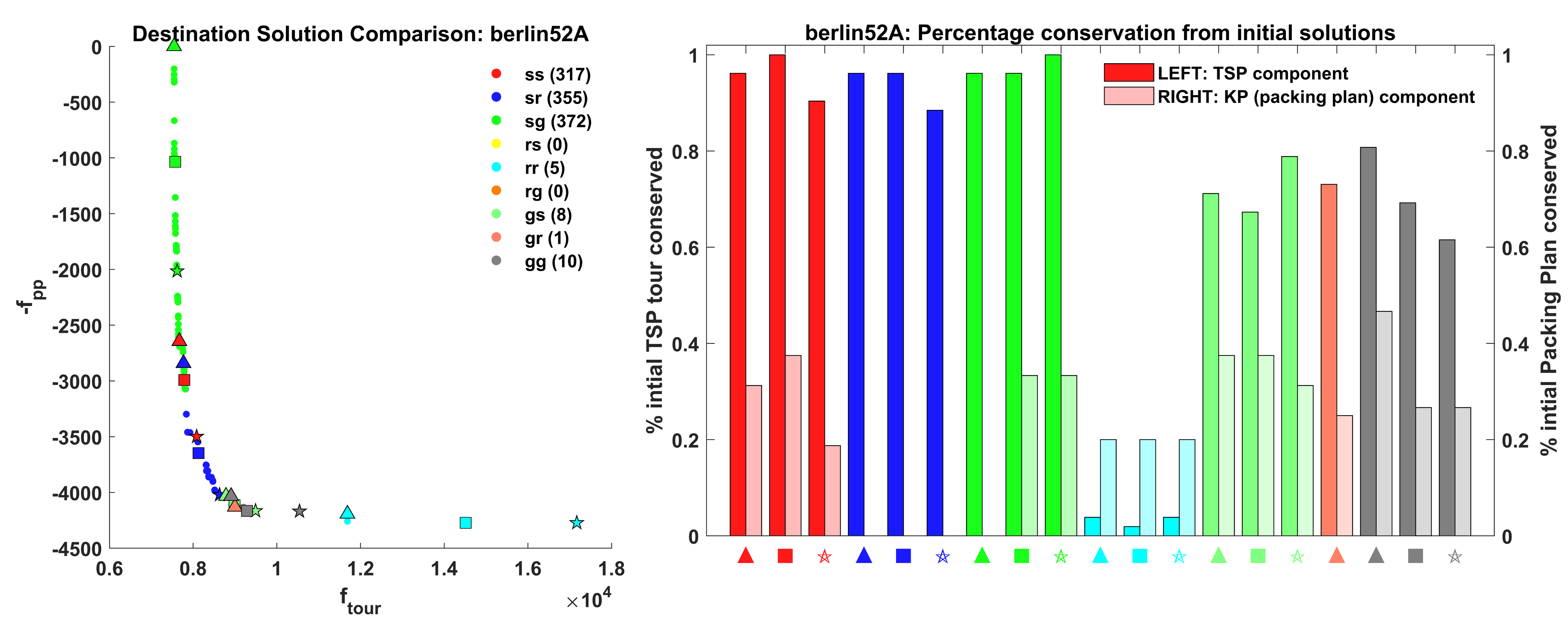}
    \caption{(left) The composite Pareto Front obtained for the berlin52A problem using the all combinations of initialization strategies with individual solutions labelled. Each labelled solution is given an index in order of increasing $f_{tour}$ value. (right) The percentage conservation between initial solution components and labelled solutions present in the composite PF.}
    \label{fig:SC_berlin52A_all_joined}
\end{figure*}

Based on these preliminary results of this localised convergence based on the different `pure' strategies, a similar composite non-dominated set is constructed using `mixed' initialization strategies (e.g. $sg$ refers to an initial set comprised of solver-based TSP components with greedily found KP components). Combination of each of the TSP and KP initialization strategies results in nine different strategies, and their contributions to the non-dominated set for the $berlin52A$ problem are shown in Fig. \ref{fig:SC_berlin52A_all_joined}.

There are a number of important observations to be made from Fig. \ref{fig:SC_berlin52A_all_joined}. Firstly, the combination of optimal TSP solutions with random packing plans ($sr$ contributed solutions) in the initial population enables different solutions to be found compared with a `pure' solver-based initialization strategy . An initial population formed of solver-based TSP solutions and greedy packing plans ($sg$) achieves many of the same solutions as the purely solver-based initialization strategy, however more non-dominated solutions are contributed overall. In terms of the percentage conservation of from initial solutions, both $sr$ and $sg$ labelled solutions maintain a similar percentage of the initial TSP-tour as the $ss$ solutions, however as might be expected, the $sr$ contributions show that none of the initial randomly-generated packing plan are retained. Given the previously unrepresented region of the non-dominated set that these $sr$ solutions occupy (\textit{cf.} Fig.\ref{fig:SC_berlin52A_pure_joined}) and this observation, this implies that a poor packing plan upon initialization gives the population more freedom to find higher profit solutions with slightly higher $f_{tour}$ values.

Similarly, the mixed greedy approaches ($gs$ and $gr$) provide novel solutions in the region of the non-dominated set with higher $f_{tour}$ values than any of the $ss/sr/sg$ methods, in the vicinity of the previously observed $gg$ solutions in Fig \ref{fig:SC_berlin52A_pure_joined}. This reinforces the notion that initialization with sub-optimal tours in the population enables greater exploration of higher profit solutions.

The mixed random approaches ($rs$ and $rg$) do not appear to provide additional representation of non-dominated solutions in the composite set, however this is likely due to the poor initial performance of all solutions with randomly generated TSP-solutions. Effectively, all methods using solver and greedy solutions to the TSP component give a head start to the initial population. This is overall towards  solutions that are minimizing $f_{tour}$; good solutions for $f_{pp}$ will require more evaluations due to the previously mentioned indirectness of a KP-solution to the fitness of a packing plan. Nevertheless, since the $rr$ response supplies the highest profit solutions of any of the methods, the random initialization strategies are retained in the further experiments.

\begin{figure*}[!t]
    \centering
    \includegraphics[width=\textwidth]{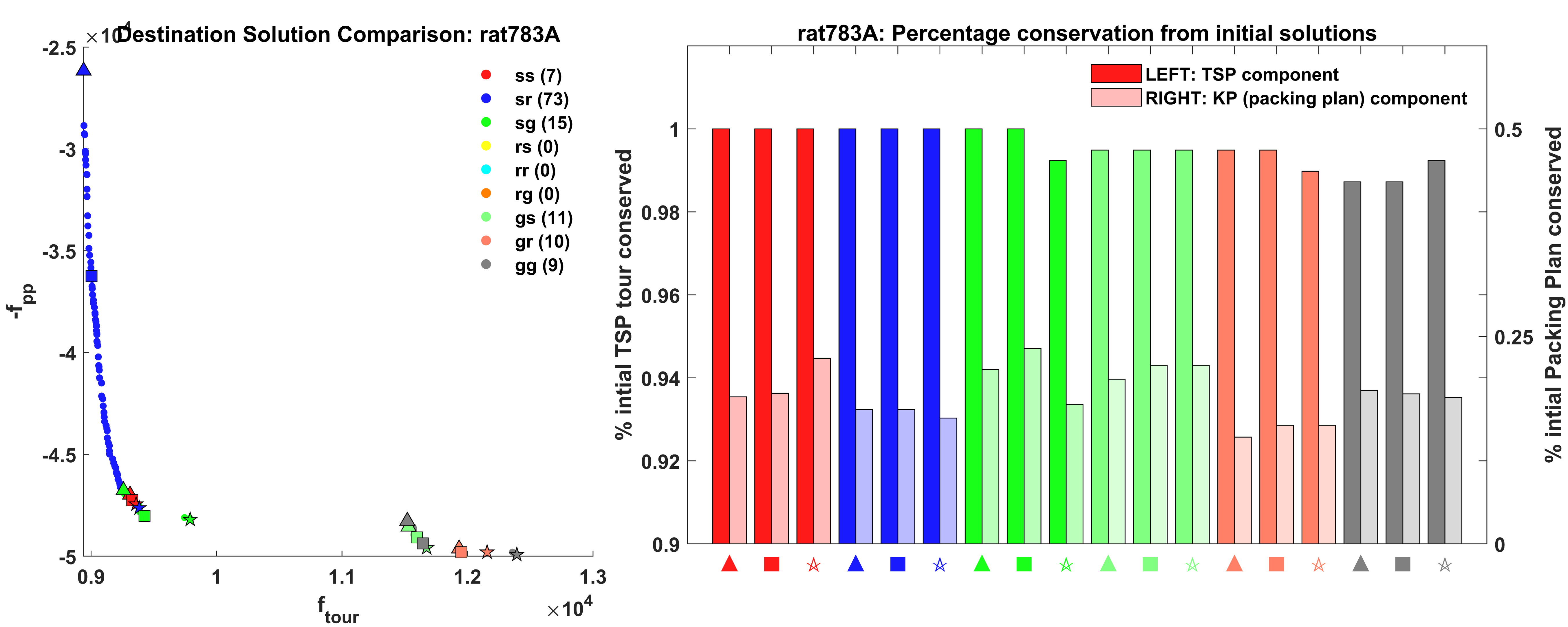}
    \caption{(left) The composite Pareto Front obtained for the rat783A problem using the all combinations of initialization strategies with individual solutions labelled. Each labelled solution is given an index in order of increasing $f_{tour}$ value. (right) The percentage conservation from initial solution components to representative solutions in the composite non-dominated set are given, colours and marker shapes correspond to those used in the left-hand plot.}
    \label{fig:SC_rat783A_all_joined}
\end{figure*}

These observations are drawn from the smallest considered problem instance for both TSP and KP components, but similar behaviours can be seen as the size of the problem increases. For example, Fig.\ref{fig:SC_rat783A_all_joined} shows the distinct groupings in the composite non-dominated set for the rat783A problem, from methods that used solver-based TSP tours in their initial populations and those that use greedily formed tours. No visible contribution is made by the strategies supplied solutions with randomly generated tours, however as mentioned, more evaluations may be required to find these highest profit solutions.

\begin{figure*}[!t]
    \centering
    \includegraphics[width=\textwidth]{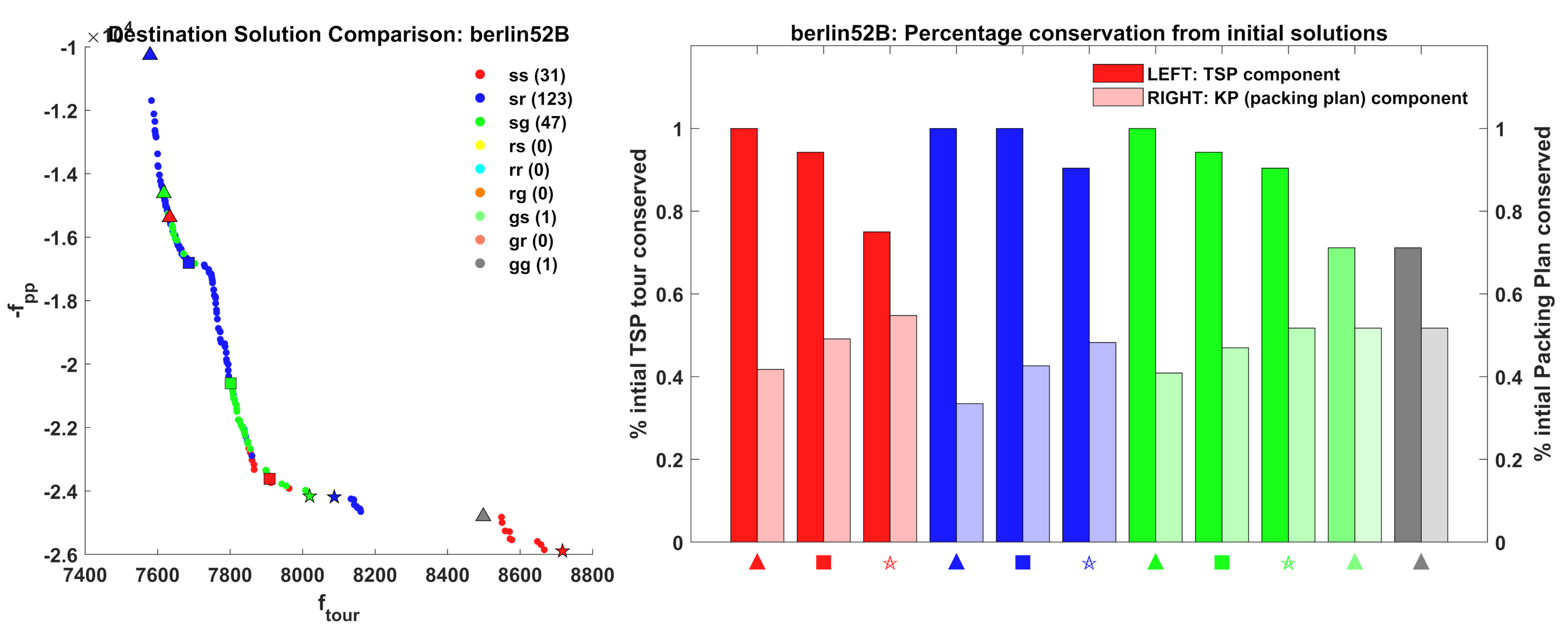}
    \caption{(left) The composite Pareto Front obtained for the rat783A problem using the all combinations of initialization strategies with individual solutions labelled. Each labelled solution is given an index in order of increasing $f_{tour}$ value. (right) The percentage conservation from initial solution components to representative solutions in the composite non-dominated set are given, colours and marker shapes correspond to those used in the left-hand plot.}
    \label{fig:SC_berlin52B_all_joined}
\end{figure*}

The differences in conservation of initial packing plans can very clearly be seen in the composite set obtained for the berlin52B problem in Fig. \ref{fig:SC_berlin52B_all_joined}. Here we see the similar grouping to the previously examined problems, however the spread of the solutions contributed by each of the methods using solvers for the TSP-component ($ss, sr, sg$) methods overlaps considerably. The labelled solutions for each of these methods are distributed such that each method's first solution is has a lower $f_{tour}$ value than any other method's second solution for example. This means that when observing the trends in the percentage conservation we see clearly that across all $s\_$ methods a lower conservation of initial TSP solution corresponds to a higher profit solution (solutions labelled with a star icon). We also see the inverse relationship with the conservation of the initial packing plans; higher conservation in higher profit solutions, even though each of the methods generates them differently. This trend implies the intuitive feature of the trade-off in this problem; for a solution to achieve a higher profit when provided with some head start to the tour component, they should retain as many items from it's initial packing plan as possible.

Ultimately, these results indicate that the chosen initialization strategy and seeding of the initial population with solutions with component fitness of different qualities has an impact on the range of solutions that can be achieved. Therefore, for the remainder of the experiments on the DTTP instances, reintialization strategies are constructed to compare the effectiveness of seeding the population with different information from solvers, greedy solutions and randomly generated solutions in responding to dynamic changes in the problem (See section \ref{sec:MethodsRein}). From the nine examined combinations from these experiments on the static problem, the eight response methods in Table \ref{tab:Algorithms} are defined.

\subsection{Responsive Solution Generation methods for the DTTP}
\begin{figure*}[!t]
    \centering
    \includegraphics[width=0.9\textwidth]{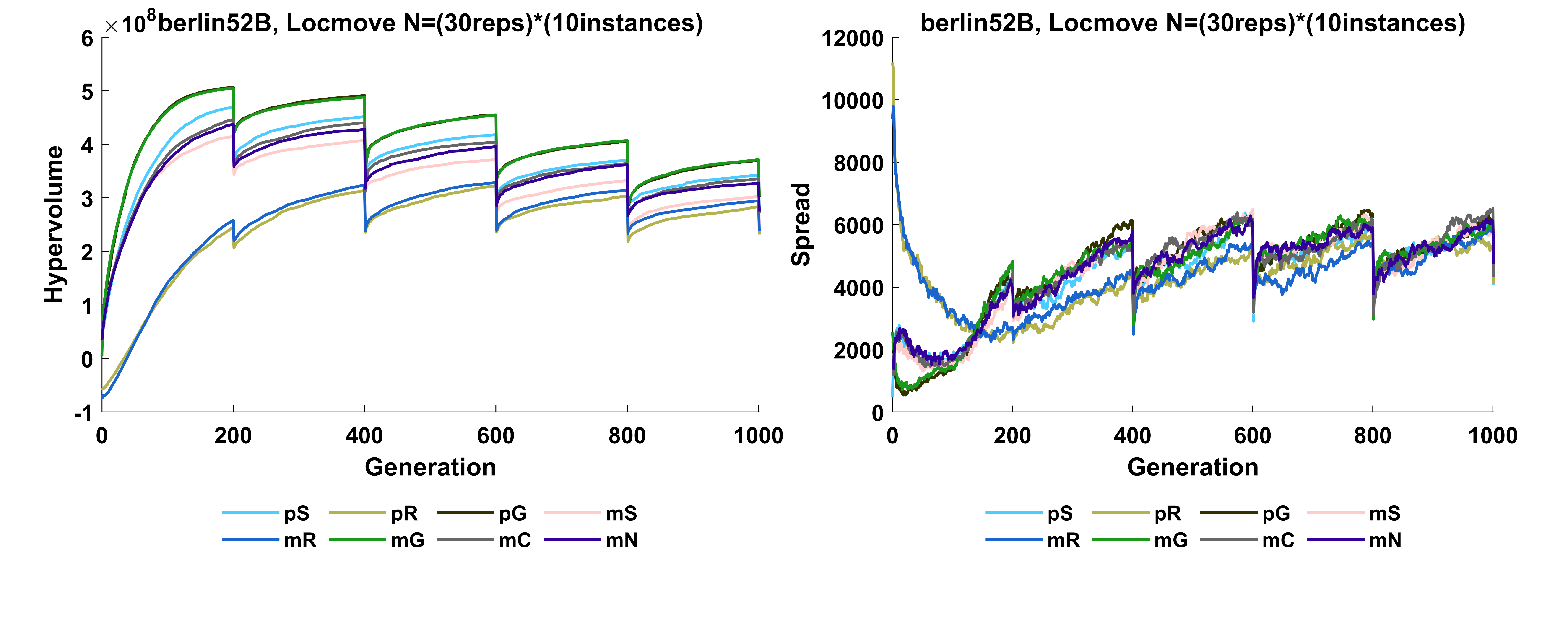}
    \caption{Hypervolume profiles for the eight strategies on the berlin52B problem with $Loc$ dynamics. Each profile is the median taken from the mean of 30 repeats on each of 10 dynamic instances.}
    \label{fig:Locmove_HV_b52B}
\end{figure*}

\begin{figure*}[!t]
    \centering
    \includegraphics[width=0.9\textwidth]{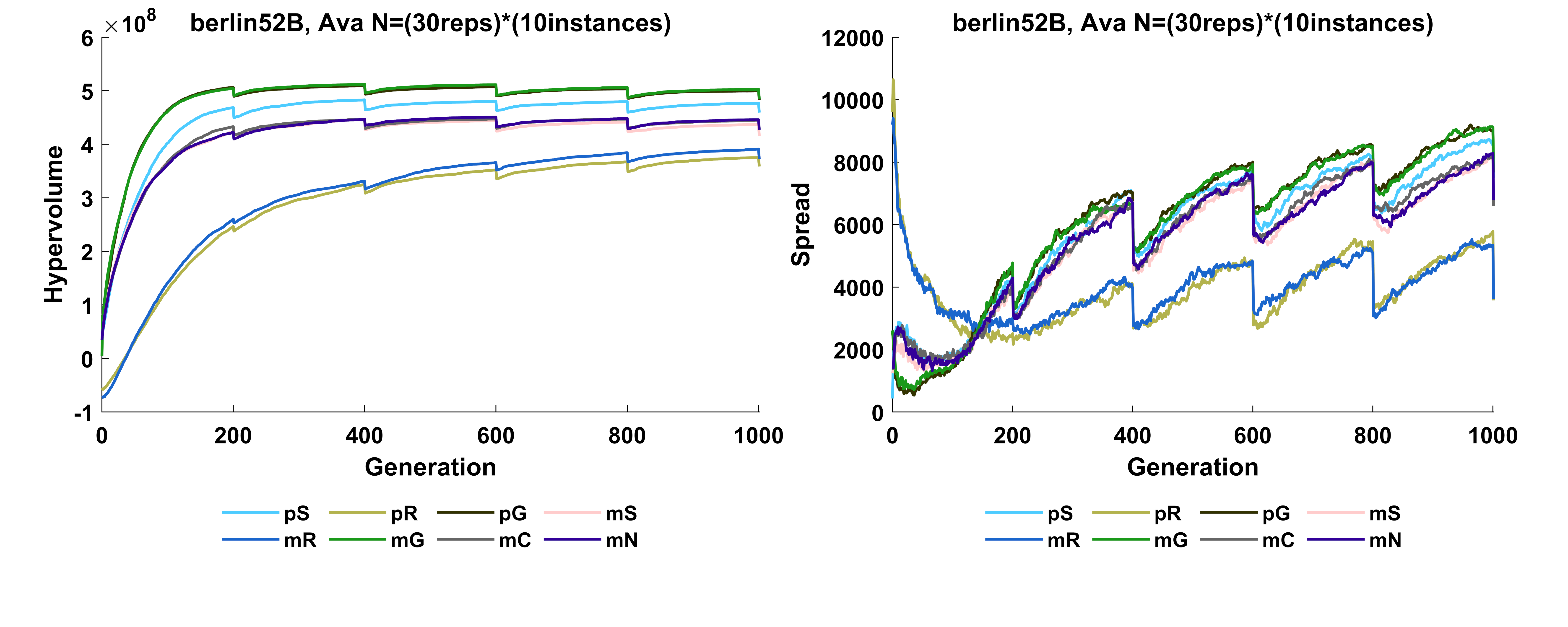}
    \caption{Hypervolume profiles for the eight strategies on the berlin52B problem with $Ava$ dynamics. Each profile is the median taken from the mean of 30 repeats on each of 10 dynamic instances.}
    \label{fig:Ava_HV_b52B}
\end{figure*}

\begin{figure*}[!t]
    \centering
    \includegraphics[width=0.9\textwidth]{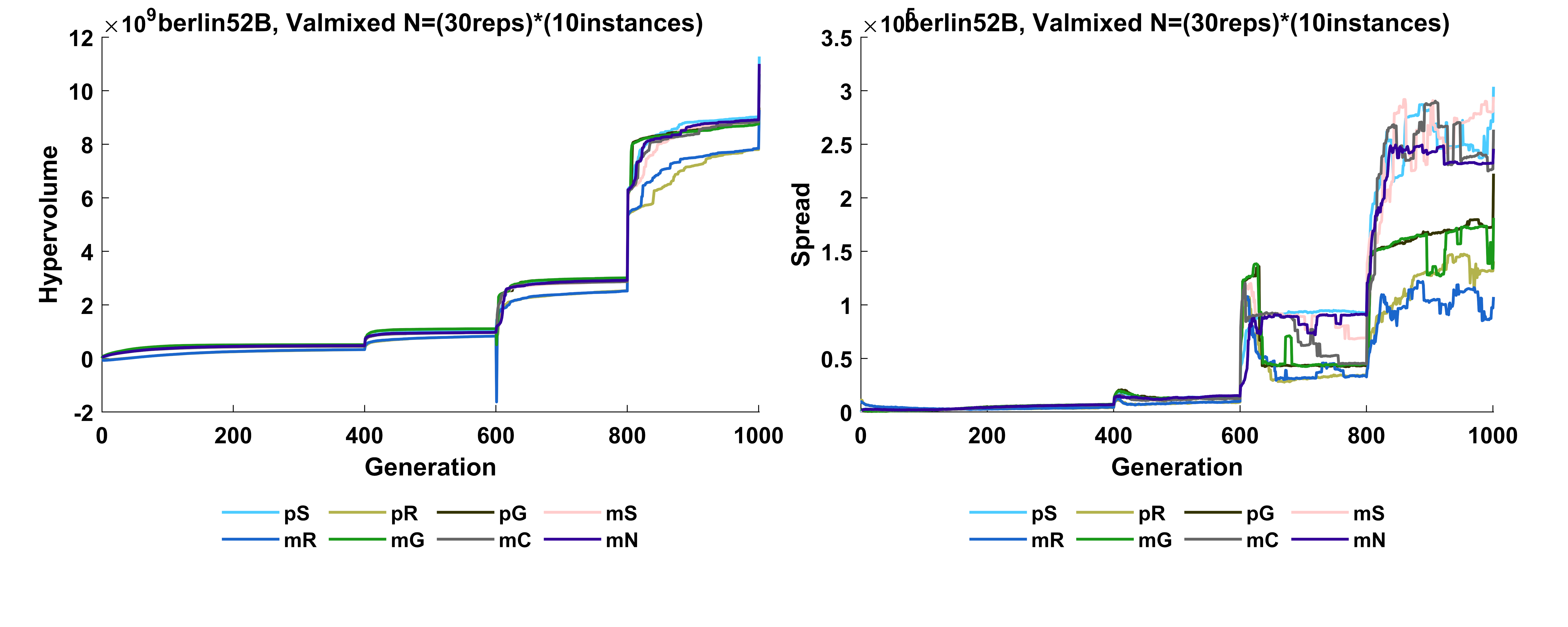}
    \caption{Hypervolume profiles for the eight strategies on the berlin52B problem with $Val$ dynamics. Each profile is the median taken from the mean of 30 repeats on each of 10 dynamic instances.}
    \label{fig:Valmixed_HV_b52B}
\end{figure*}

The mean hypervolume and mean spread measure profiles in figures \ref{fig:Locmove_HV_b52B},\ref{fig:Ava_HV_b52B} \& \ref{fig:Valmixed_HV_b52B} are shown for the $berlin52B$ problem for each of the three types of dynamic changes $Loc$ (A \& B), $Ava$ (C \& D), and $Val$ (E \& F). Each profile is the result is the mean of 30 repeats for each of 10 patterns of dynamic changes. Each examined method is given the same set of patterns of dynamic changes with which it must provide the responsively generated solutions to in order to make their performance comparable. The noisy character of the spread measurement profiles is to be expected; replacement according to the domination criteria do not maximise spread and therefore decreases in the size of the hyperbox between extreme solutions may decrease as better solutions are found that dominate these previous solutions.

\subsubsection{Observable differences of impacts between the different types of dynamics}
In both the hypervolume (left) and spread (right) profiles, each of the 5 changes has a visible impact on the performance of the solution set in terms of both the achieved hypervolume and spread. It is important to note that changes in all three types of dynamic scenarios affect the maximum achievable hypervolume; this is intuitive for $Loc$ and $Val$ dynamics since the minimum distance tour and maximum achievable profit are altered respectively. Under the $Ava$ dynamics however, since it is only items’ city assignments indexes that change, the effect of this change is more complex than the direct changes in the other components.

Whilst the impacts of transition to each new dynamic interval appears metered and straightforward for the $Loc$ and $Ava$ dynamics, there are drastic changes in the measurement profiles for the methods on $Val$ problems. Whilst the magnitude of the changes is fixed, they persist such that the resemblance between the problem states and the initial problem state decreases with successive dynamic intervals. This appears to be most significant for the problems with $Val$ dynamics in terms of both the achievable spread and hypervolume values as they become much greater as the number of total generations and therefore, number of change events increases. This is an indication that altering the values of the items in the problem definition (both as a mixture of increases and decreases) can result in large differences in the extent of non-dominated set in the objective space; implying that small changes from the initial problem set can result in both greatly altered achievable solutions and clear separation of response method performance. Successive changes in this component of the problem therefore make it more difficult for some response methods to mitigate the impacts of changes and enable finding competitive solutions in the new dynamic interval. 
Another important feature of these measurement profiles is in the magnitude of the post change decrease and its amelioration through the introduction of different solutions into the population immediately after a change. Also, the ability to find good solutions (that achieve high hypervolume and spread measurements) is reliant on the composition of the response solutions, illustrated by the non-convergence and dissimilarity of achievement by each of the methods.

\subsubsection{Inferences on suitability of different methods for different types of dynamics}
Together the two measurements give an indication of the quality of the population of solutions achieved by the different methods and because of the preliminary observations, give an indication of the performance with respect to convergence and extent in each of the objectives. We know from the observed localization of the non-dominated solution sets when using different initialization methods that many solutions can often be found with shorter length tours and comparatively fewer high-profit solutions are found. The propensity of a method for finding such solutions depends on the aggressiveness of the exploitation/concentration on the optimality of the tour component of a solution; greedily and randomly generated tours are longer upon initialization and as such have more freedom to achieve higher profits.
For some problems, as highlighted in these profiles, the nature of the problem components suits a particular method or methods more than others. For example, the method which responsively introduces solutions with both greedily generated tours and packing plans (pG) achieves both the highest spread and hypervolume values during each of the dynamic intervals for this problem with $Loc$ and $Ava$ dynamics. The combination of achievement in both measurements tells us that this method can both find good solutions in the knee-point of the non-dominated set (high HV) and that the extent of solutions with higher profits (and longer tours) is better (high spread) is than the other methods. 

\subsubsection{Justifications for using ranking to analyse relative performance of the different response mechanisms.}
\begin{figure*}[!t]
    \centering
    \includegraphics[width=\textwidth]{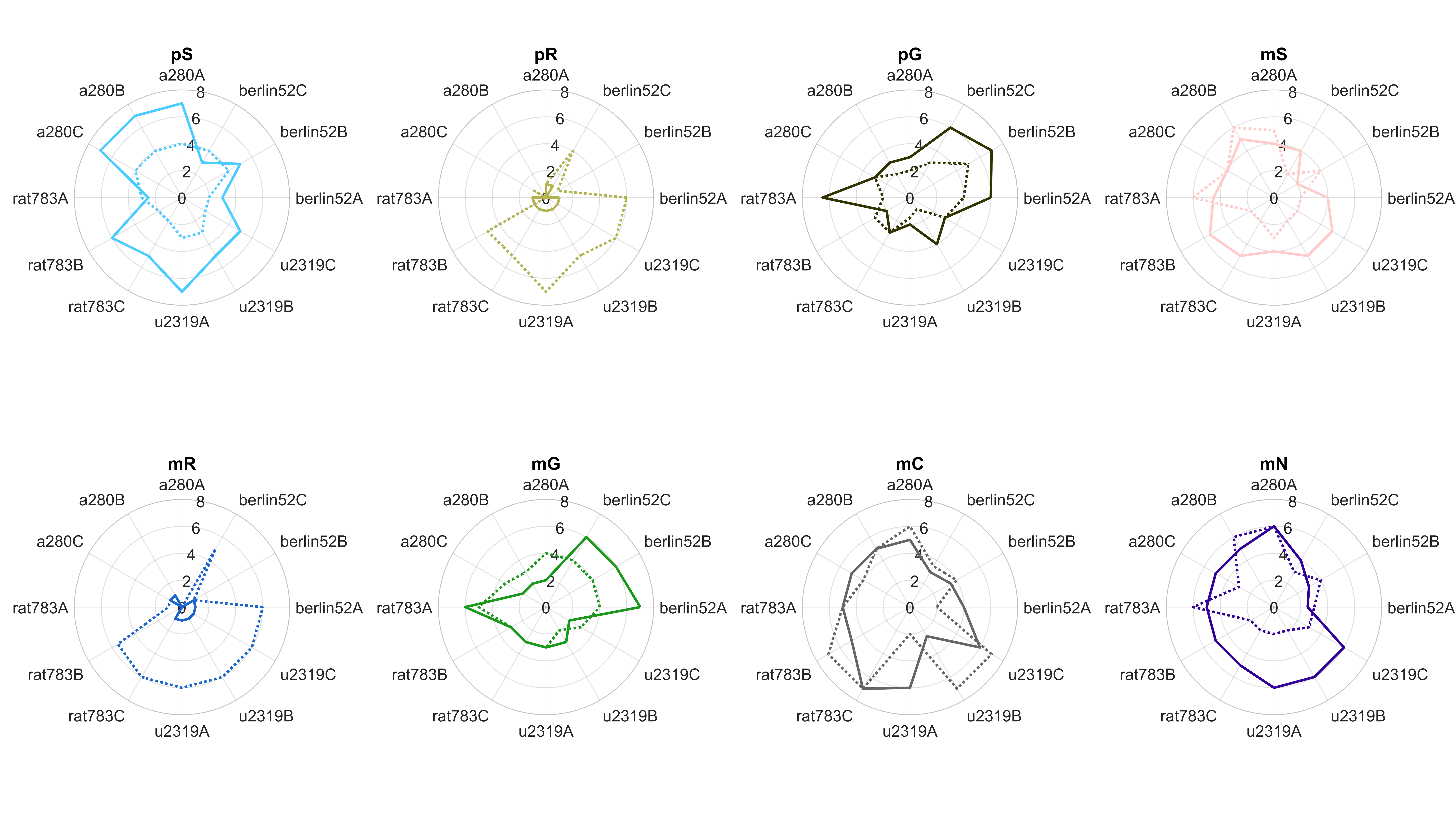}
    \caption{Polar plots of median end-of-interval hypervolume ranks of the different strategies on the DTTP with city location dynamics. Each subplot corresponds to one of the response methods listed in Table \ref{tab:Algorithms}. Solid lines are hypervolume rankings, dashed lines are spread rankings.}
    \label{fig:LocmoveRanks}
\end{figure*}

\begin{figure*}[!t]
    \centering
    \includegraphics[width=\textwidth]{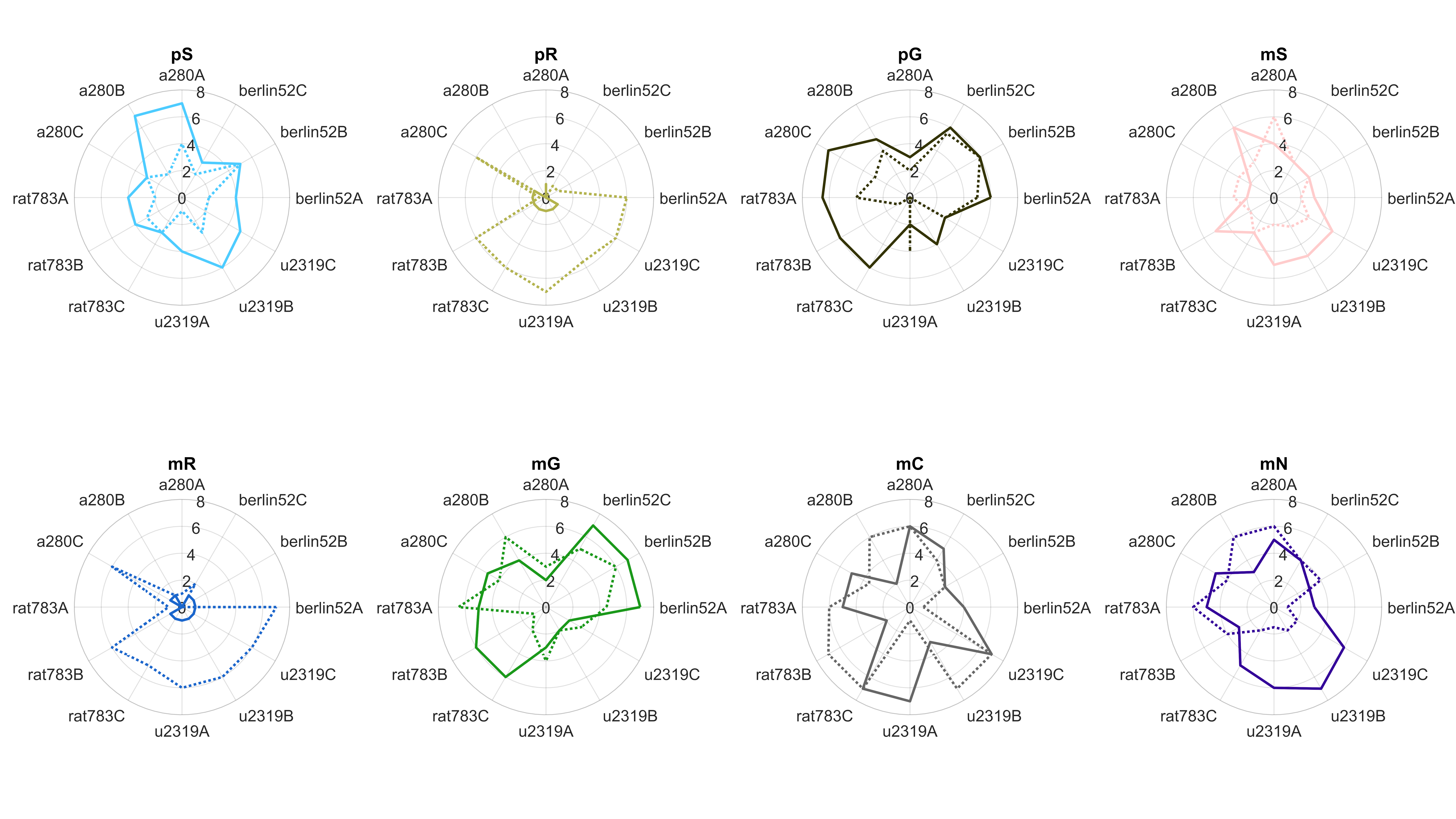}
    \caption{Polar plots of median end-of-interval hypervolume of the different strategies on the DTTP with item availability dynamics. Each subplot corresponds to one of the response methods listed in Table \ref{tab:Algorithms}. Solid lines are hypervolume rankings, dashed lines are spread rankings.}
    \label{fig:AvaRanks}
\end{figure*}

\begin{figure*}[!t]
    \centering
    \includegraphics[width=\textwidth]{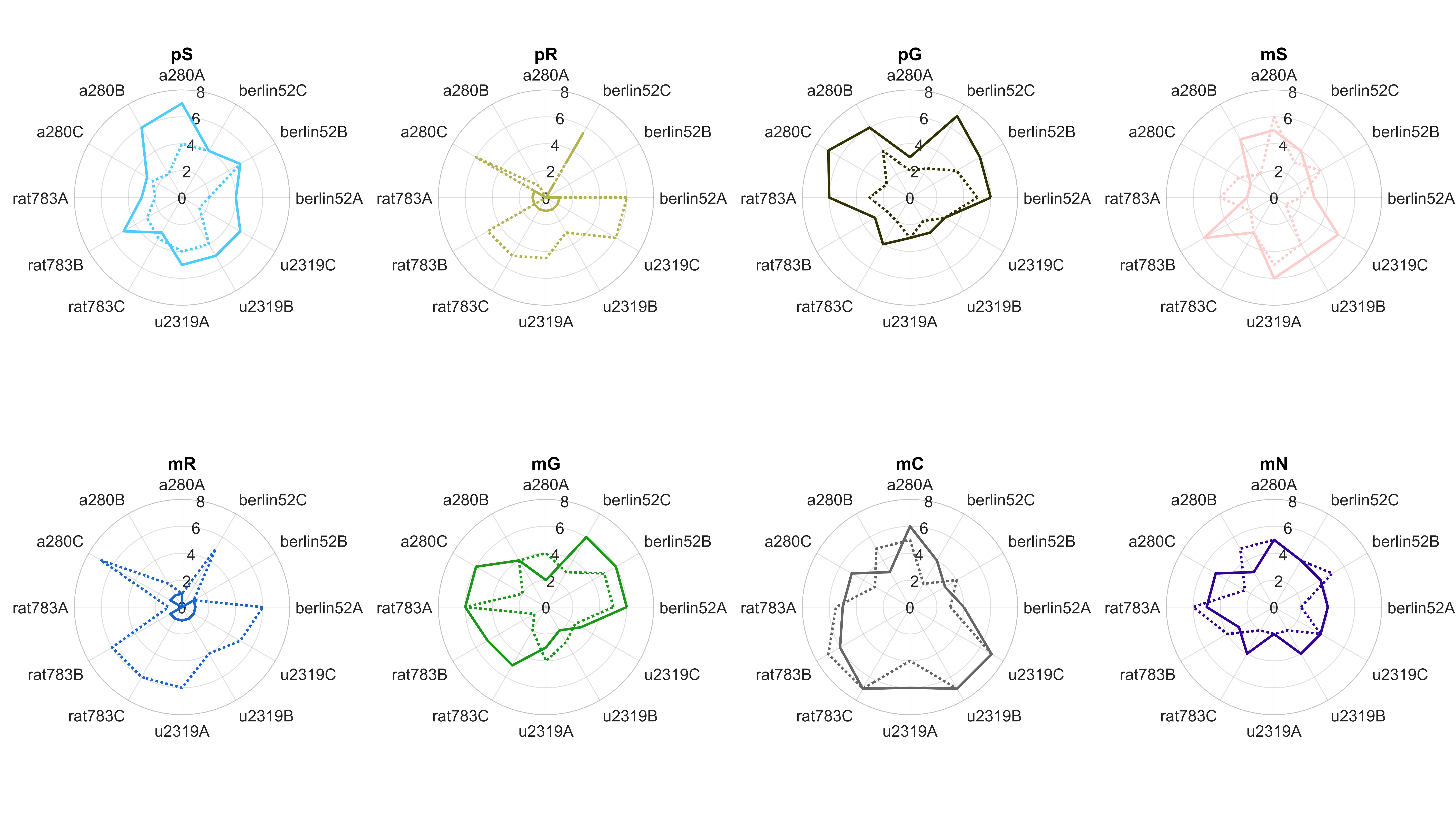}
    \caption{Polar plots of median end-of-interval hypervolume of the different strategies on the DTTP with item value dynamics. Each subplot corresponds to one of the response methods listed in Table \ref{tab:Algorithms}. Solid lines are hypervolume rankings, dashed lines are spread rankings.}
    \label{fig:ValmixedRanks}
\end{figure*}

There are several factors to consider in order to allow for coherent analysis and comparison across the intervals in each problem and across the range of problems. These include previous statements on the disconnect between successive dynamic intervals and the observation that the maximum achievable hypervolume is not consistent between intervals. Similarly, the theoretical maximum hypervolume for each of the problems is dependent on both the size of the TSP and KP components in the problem definition. Therefore, to enable meaningful comparison of the performance of each method in response to changes, we propose to rank the methods by their achieved end-of-interval (pre-change) hypervolume and spread values. This represents the performance of a method in the new dynamic environment after having been supplemented at the generation-of-change with solutions generated according to each method’s construction mechanism. The median end-of-interval rankings are aggregated across the dynamic instances and are reported separately for each type of dynamics in figures \ref{fig:LocmoveRanks}, \ref{fig:AvaRanks} \& \ref{fig:ValmixedRanks}.




    \subsubsection{Interpreting ranks and comparative performance of method in $Loc$ dynamic problems}
As described previously, the \textit{Loc} dynamics refer to change in the coordinates of city locations within a feasible translation area. For the tested problems, at the start of each dynamic interval, the change was applied to only two cities ($d_{N}=2$), regardless of the total number of cities in the TSP component. Figure \ref{fig:LocmoveRanks} shows the ranking for the end-of-interval hypervolume (HV) and spread metric measurements across the different DTTP problems with \textit{Loc} dynamics. The labels corresponding to the different methods are explained in Table \ref{tab:Algorithms}. 
As expected, the random methods, pR and mR achieve the lowest HV ranks across the problems showing they are not competitive with any of the other methods in terms of responding to a dynamic change. Interestingly, even the passive mN approach outranks the random methods in achieved HV for every problem. However, as noted in the previous comparison of contributions to the non-dominated set in the static problem, the diversity introduced by these methods is useful for exploration of the search space and the achievement of a diverse solution set. This is somewhat illustrated by the high spread ranks achieved for the larger TSP-component problems by pR and mR; evolution of the population occurs towards a number of well spread solutions being found by these more explorative methods but do not achieve competitive hypervolumes. In general, the random based methods do not provide competitive solutions compared with the other methods and are ineffective as a response mechanism in these problems.

In contrast, the solver based methods of pS and mS show interesting trends across the different problems with \textit{Loc} dynamics. The pS method achieves the highest hypervolume ranks on all of the \textit{a280A, B} \& \textit{C} and \textit{u2319A} problems, with good ranks on all other problems with the exception of \textit{berlin52A} \& \textit{C} and the \textit{rat783A} problems. Across all problems however, the spread ranks are poor compared with the other methods. Similar observations can be made from the mS method, however lower HV ranks and higher spread ranks are present, implying a compromise of the positives and negatives from the pure solver-based method by introducing mixed-origin packing plan solutions.

These high HV ranks and low spread ranks imply a good convergence for some portion of the non-dominated set but a limited diversity compared with other methods. As noted in the previous comparison of contributions to the non-dominated set, the solver-based methods favoured finding solutions in the minimum-tour region of the objective space, neglecting coverage of higher-profit solutions. This would explain the reduced spread rankings observed in these methods as coverage of the non-dominated set of solutions is mostly limited to solutions with shorter tours.

For the greedy methods, it was previously noted that the introduction of greedily-constructed solutions enabled achievement of higher profit solutions with slightly longer tours, and here there is almost an inverse performance ranking to the solver-based methods. Best performance is achieved for pG and mG across the \textit{berlin52A, B} \& \textit{C} and \textit{rat783A} problems, where both methods achieve high HV ranks, but relatively lower spread ranks. This similarly reflects the limitation of diversity in the population as with the solver-based methods; mG intuitively achieves slightly higher spread ranks on all problems, since the packing plan components of the response solutions comprise random and solver-based generation in addition to the greedy construction method.

From considering the results of the pG/mG and pS/mS methods, it seems that particularly for the problems with the $a280$ and $berlin52$ TSP-components, that the impact on the difficulty of the problem that the nature of the packing plan component has is independent of the TSP-component. the pG/mG and pS/mS methods respectively achieve high HV and lower spread ranks on the two sets of problems with the $berlin52$ and $a280$ TSP-components. This illustrates the limitation of using a single initialization and solution generation method as it implies minimum tour solutions may be easy found by these methods on each problem, but the lower spread measurement indicates that coverage of higher profit solutions is poor. 
This leads us to the consideration of the combined approach in the mC method. Here a combination of solution components is used that covers all combinations of solution generation methods; with tours and packing plans from solver, greedy and random origins are combined together in all possible combinations (without increasing the population size). Generally, the rankings show a decent all-round performance in terms of HV and spread. The mC method achieves mid-range values across the smaller TSP-component problems (on which the pS/mS and pG/mG methods appear to consistently achieve the best hypervolume ranks of any method). However, whilst these other methods don’t consistently preserve their performance as the scale of the TSP and therefore KP components increase, the mC methods still achieves some of the best HV and spread ranks on these `larger’ problems. 
In some cases, such as $u2319A$ \& $B$ and for $berlin52A$, the rankings for spread and HV are not consistent with each other. In each of these cases the ranks are better overall than many of the other methods, but it is not immediately intuitive as to why there is such an extreme discrepancy in the rankings achieved on these specific problems. Generally however, the performance of the mC method supersedes many of the other methods, particularly on the problems with more cities. 

There are interesting trends in the rankings achieved by the passive method, mN, within which no newly generated solutions are added into the population in response to a dynamic change. The hypervolume ranks increase with the size of the TSP and KP components in the problems, however the spread ranks achieved do not increase beyond the problems with the mid-sized $a280$ \& beyond the $rat783A$ problem. Good rankings for both HV and spread (on the smaller TSP-component problems) imply that the magnitude of change is not so large that pre-change solutions are still competitive (in terms of a small spread of good solutions being robust) for the post-change problem environment. It reflects the comparative ease of finding shorter-tour solutions with minimal profit values, compared to finding non-dominated, longer, high-profit tours.

This trend does not continue as the size of the TSP and KP components increases, implying that despite the magnitude of the changes $d_{N}=2$ being constant across the problems, the total number of cities in the problem plays a considerable role in determining the effectiveness of different response methods. Moreover, whilst the hypervolume ranks remain high for the passive approach, the decreasing spread ranks with increasing problem size (both TSP and KP components) further illustrate the disparity between the ease of finding shorter low-profit tours and longer, high-profit tours. 

Despite the nature of the dynamics being isolated to a change in the city locations, in these cases, the high HV and low spread imply a coverage of the minimum-tour region and the knee-point of the non-dominated set, with little to no consideration of the high-profit region of objective space. From the preliminary solution conservation results, it becomes clear that unless these high profit solutions can be deliberately sought out by introducing diversity in the population, as the tendency of the population is to focus on minimum-tour solutions. Therefore an effective response mechanism can introduce solutions that will lead to both high-profit and minimum tour solutions, even for these problems with a TSP-component size towards the top of the applicable range for EAs.

The results for this type of dynamics ultimately indicate that when the TSP-component is small, $a280$ \& $berlin52$, the nature and nuance of each of these TSP problems controls whether solver based methods or greedy based methods are the best at achieving solutions with shorter tours in the non-dominated set. The number of items and their profit-weight correlation is largely ignored by these methods, since their populations become saturated with near-minimum-tour solutions with little coverage towards the maximum profit solutions ( highlighted in the preliminary results for the static problem) as reflected by the consistency of the HV ranks and their disparity to the achieved spread rankings on these problems. It becomes apparent that as the size of the TSP-component increases, the performance in terms of similarity in HV and spread rankings remains poor. Only with the combined approach of mC are the rankings consistent with each other on some of the problems with larger TSP components. Ultimately, despite the discrepancies for the $u2319A$ and $u2319B$ problems, the mechanism of complete heterogeneity in the responsively introduced solutions is broadly suitable across the range of examined problems with $Loc$ dynamics. 

\subsubsection{Interpretation of rankings for problems with $Ava$ dynamics}
Consistent with the observed performance on the problems with $Loc$ dynamics, the pR \& mR methods do not provide competitive rankings for HV on any of the problems, and therefore the disparity in HV rankings observed for the larger problems is indicative of poor performance of the solutions obtained by these methods. The introduction of randomly generated solutions is in fact detrimental to the algorithms overall performance since the ranks achieved are worse than the passive approach of mN.
Whereas the solver-based methods in pS \& mS widely achieved high HV ranks (but with poor spread rankings) across many of the tested problems where $Loc$ dynamics were occurring, the general performance across these problems is decreased when $Ava$ dynamics are the changing aspect of the problem. Conversely, the range of problems under which the greedy methods pG \& mG can provide high HV rankings is increased between these types of dynamics. Overall however, the introduction of greedily constructed and solver-based solutions post-change results in poor diversity in the non-dominated set, as previously noted. The effects of this however are less severe when using the greedy method pG as the spread rankings are consistent with HV ranks across the $berlin52$ problems. Furthermore, the mG approach, which combines greedily constructed tours with packing plans from a mixture of generation methods, maintains or improves the spread rankings across the problems further.

Whilst the introduction of solver-based and greedily-constructed solutions after a dynamic change event can lead to good coverage of the minimum-tour region of the non-dominated set, the diversity introduced by a mixed approach to packing plan solution construction for these response populations enable a better solution coverage without loss of hypervolume rankings. Interestingly, where this extra diversity assisted in `rounding-out' the performance of the solver-based method (as mS) in the problems with $Loc$ dynamics, here under $Ava$ dynamics the spread ranking performance across the problems is slightly improved but mostly different from the pS rankings. Generally, the pS \& mS methods perform worse for problems with $Ava$ dynamics than they did for problems with $Loc$ dynamics.

The combined approach, mC, again shows achievement of good rankings across the widest range of the examined problems but struggles with problems with the $B$-type KP component (5 items per city, similar item weights/profits) but not $C$. One potential justification for this could be that, similar item weights (as shown in Fig \ref{fig:ABCitemdists}) may mean that the potential impact of the changes is lesser in terms of difference between subsequent problem states, and as such the solver-based and passive approaches allow for better performance than the introduction of more diverse solutions in the combined methods.
Inconsistencies in HV and spread rankings noted before for $u2319A$ \& $u2319B$ are present in in Fig \ref{fig:AvaRanks} for the $Ava$ problems too, implying some specific characteristic in these problems that is controlling performance beyond the impact from the type of dynamics. It may be that the previous justification for the performance on the $B$-type problems is also the cause of this discrepancy and in the $Loc$ problems; the ranks on the problems with smaller TSP-components may be inflated due to the relatively mediocre performance of other methods generally across these problems.
The same trends are not present in the $Val$ problems, but since it is the values of the items that change in these problems, this phenomenon appears to be prevented such that a combined approach performs better for problems with this type of dynamics.

\subsubsection{Interpretation of rankings for problems with $Val$ dynamics}
The respective trends observed in the $Loc$ and $Ava$ problems for pR/mR , pS/mS and pG/mG mostly persist here. This includes pR \& mR methods being unsuitable for all problems in terms of providing useful solutions in response to dynamic changes. Similarly, the solver-based methods pS \& mS maintain decent HV ranks on $a280A$ \& $B$, (with the best HV ranks for $a280C$ achieved by pG here) and the greedy method pG achieving high HV ranks across the $berlin52$ problems too. 
Interestingly, the mS method has better spread rankings compared with pS, and likewise, mG retains the previous seen HV rankings on the $rat783$ problems with better spread rankings than pG on most problems. This implies that where the item values change in the problem, the increase in the diversity of the introduced packing plans is useful in enabling good solutions to be found in the next interval.  The higher spread indicates that this includes some higher-profit solutions.
The combined method, mC appears to achieve high ranks for the $rat783$ \& $u2319$ TSP-component problems across all $A, B$ \& $C$-type problems. Just as observed for the other methods, there are mid-range rankings for this method on the smaller TSP-component problems. As before, one explanation for this is that the size of these problems is sufficiently smaller that solver and greedy response methods can perform well enough in terms of generating useful solutions after a change.  This means that as the size of the problem increases, the effectiveness of these highly exploitative and relatively non-diverse methods is limited. Increasing the magnitude of the dynamic changes in further experimentation may more clearly illustrate the differences in performance between the response methods.
Interestingly the HV rankings for the passive approach, mN, on problems with $Val$ dynamics are not consistent with those observed for $Loc$ and $Ava$ problems. Whereas previously we saw an increase in the achieved HV rank with increasing TSP-component size (in part due to the above justification of other methods declining performance), under these dynamic conditions, there is no such trend. The achieved ranks for this method remain in the midrange across all the problems, highlighting both that some response methods prove detrimental and that there are better ranks consistently achieved by other methods (mostly the mC method).

\subsubsection{Discussion}
The introduction of dynamics to these three parts of the TTP formulation provide different levels of difficulty in terms of the resulting problem and the methods which can continue to find good solutions despite the changes.

This becomes clearer when there are clearly observable trends in the methods which perform well; the effect of the dynamics is consistent across different problems and instances of the dynamics such that the same methods can perform well consistently.

In contrast, for problems where no clear optimal performance is visible across the applied methods, this implies that the changes introduced into the problem do not allow for consistent success by a particular dynamic response method. We can therefore say that introducing this type of dynamic changes in the problem generates a problem environment that is more difficult to respond to with the examined population seeding methods; harder problems are generated.
For example, $a280B$ has no method that achieves consistently the highest ranks in both spread and hypervolume across all three types of dynamics. This implies that no single method is the most suitable for this problem under every instance of the dynamic changes.

In the combined approach there is an expectation that the introduction of randomly generated solutions may lead to an improved coverage of solutions in the high-profit region of the non-dominated set, based on the preliminary observations. However a combination of the controlled population size and the observed bias for coverage of the minimum-tour solutions means that upon introduction, rank-based and crowding-distance based replacement is likely to discard solutions that would enable coverage of high-profit solutions in future generations.
As described, this also explains why the relative performance of the mC method appears to be worse on the problems with smaller TSP-components across all three types of dynamics. Another explanation may be in the relative coverage achievable within an introduced population with respect to the number of solutions-per-construction-method. Since mC uses the same population size but divided based on the six other generation methods, the total number of solutions generated by each method (and therefore the information exploitation) may be less than for the other response methods.

The other methods are capable of competitive performance in terms of the spread and hypervolume achievement because the size of the problem is limited, and in the case of the problems with $berlin52$ component, well suited to solving using greedy approaches. Conversely, the performance rankings of these methods are consistently poor as the size of the TSP component and therefore of the KP component, of the problem increase; verifying the merit of the combined approach.
Therefore because of these observations, future work should address the universal out-performance of the solver and greedy based methods for these problems with smaller TSP-components. Since the best response method appears inconsistent across the different TSP-components and KP configurations in the problem, there is scope to improve the cohesiveness of the information from each method within the combined response. to allow good performance . These improvements are likely to improve overall performance on all of these problems including for larger-TSP problems as well.

One method for this could be an island-model approach or cooperative co-evolution methodology\cite{Mei2016} that, based on the observed solution localization, enables isolated development of solutions constructed through different generative methods. A composite or mixing population can also be maintained in parallel to these. This would enable a persistence of the high-profit solutions that are otherwise rapidly replaced by minimum tour solutions in the current methodology and provides an intuitive methodology with which to address the observed imbalance in solution convergence.

Similarly, an important extension to this work comes in considering the multiple occurrences of dynamic aspects in the problem. Here we limited the dynamic changes within the problem to be of one type, either city location, item availability mapping or item value, during the problem. Consideration of combinations of these furthers the context-driven formulation of realistic problem scenarios under this framework; it follows naturally that a change in depot location (city location dynamics) may be accompanied by changes in the values associated with the items it contains. Development of the realistic problem features, such as these dynamics within the TTP enables formulation of problems that more closely replicate real-world scenarios. Furthermore, the richness of the dynamic characteristics, including the interaction and coincidence are important to consider in future. Limited work exists that considers the operation of dynamics on different frequencies \cite{Herring2019}, and no such consideration exists in multi-objective combinatorial domains.

It is clear from the results obtained for the dynamic TTP problems, that a randomisation based initialization method is does not provide competitive results compared with the others tested. 
Solver-based initialization and re-initialization (in response to changes in the DTTP) provided better hypervolume values in some of the examined cases, however performance appears to vary depending on the TSP component of the problem. 
There is value in some problems to employing greedily initialized solutions initially and in response to changes, in line with Wagner's observations that longer tours are required for better TTP solutions. 

Since the TSP initialization method resulted in greater fluctuations in performance than the KP initialization method, there are mixed results between pure, mixed and combinations methods. 
In many cases, the incorporation of diversity in the packing plan was beneficial for increasing the spread of solutions achieved. 
Combined approach to reinitialization offers reasonable performance across all problems, and achieves consistently high ranks for the largest of the problems examined, where the other response methods do not consistently perform well. 
Between the different types of dynamics, we see similar trends in performance for $Loc$ and $Ava$ dynamics and quite different trends on $Val$ dynamics. The compound effect of successive changes to the problem appear to have a greater effect on problem difficulty in the case of $Val$ dynamics.

There is potential here to employ co-evolutionary tactics (similar to \cite{Mei2016}) which may offer improvement on the current combined approach as solutions from different initialization mechanisms would not directly compete with one another. This would allow for extensive isolated development for higher profit and longer tours that otherwise would be rapidly dominated and replaced by solver-based near-minimum $f_{tour}$ solutions.

From these results, we see that a passive approach provides competitive hypervolume rankings on some of the problems, increasingly so as the size of the TSP component increases.
This highlights that the range of parameters controlling the magnitude of the dynamic changes should be further investigated to obtain the maximum operational range of strategies with respect to the dynamic version of the TTP. To further the pursuit of realistic formulation of optimisation scenarios, the interaction and co-occurrence of the defined dynamic properties should also be considered. These points, together with the study of the effects of change frequency and the synchronicity of the dynamics are topics reserved for further investigation.

\printbibliography

\end{document}